\documentclass[10pt,twocolumn,letterpaper]{article}

\usepackage[pagenumbers]{cvpr} 

\usepackage{tikz}

\def\figmethodpath{./fig/method/}
\def\figexperimentpath{./fig/experiments/}

\graphicspath{{\figmethodpath}, {\figexperimentpath}}

\newcommand{\figmethod}[1]{\figmethodpath#1}
\newcommand{\figexperiment}[1]{\figexperimentpath#1}

\newcommand{\rowvec}[2]{\ensuremath{\left(#1~#2\right)}}

\newcommand{\vecop}[1]{\ensuremath{\text{vec}\left(#1\right)}}

\renewcommand{\vec}[1]{\mathbf{#1}}

\newcommand{\vectheta}{\mathbf{\theta}}

\definecolor{cvprblue}{rgb}{0.21,0.49,0.74}
\usepackage[pagebackref,breaklinks,colorlinks,allcolors=cvprblue]{hyperref}

\def\paperID{} 
\def\confName{CVPR}
\def\confYear{2026}

\title{Towards Reconstructing and Recognizing Objects\\behind Fragmented Occlusion using Video}

\author{Roman Pflugfelder\\
Austrian Institute of Technology\\
Digital Safety \& Security Department\\
Technical University of Munich\\
School of Computation, Information and Technology\\
Munich Center for Machine Learning (MCML)\\
{\tt\small roman.pflugfelder@\{ait.ac.at|tum.de\}}
\and
Michael Lindenbaum\\
Technion\\
Faculty of Computer Science\\
{\tt\small mic@cs.technion.ac.il}
\and
Daniel Cremers\\
Technical University of Munich\\
School of Computation, Information and Technology\\
Munich Center for Machine Learning (MCML)\\
{\tt\small cremers@tum.de}
}

\begin{document}
\maketitle
\begin{abstract}
Perceiving objects through fragmented occlusions, such as foliage, is a natural capability of the human visual system but remains a fundamental challenge for computer vision. This work addresses the problem of object reconstruction, or de-occlusion, in the presence of high-density, fragmented occlusions. We propose a reconstruction framework that operates on a sequence of grayscale images of a moving object, acquired by a single camera, either stationary or in motion. The algorithm estimates the object’s relative positions across frames by jointly registering the images and optimizing the alignment parameters to maximize the contrast of their average, referred to as the integrated image. This integrated image serves as the reconstructed representation of the object.
Experimental evaluations on both synthetic and real video sequences demonstrate the feasibility of accurate reconstruction under occlusion densities exceeding 90\,\%. 
The proposed approach is competitive with synthetic aperture techniques, which are limited to static objects and require complex camera setup with careful calibration. Furthermore, we show that the integrated image enables reliable object classification under severe occlusion when used as input to standard, off-the-shelf classifiers.
\end{abstract}

\input{\figmethod{reconstruction}}

\section{Introduction} \label{sec:intro}
Fragmented occlusion occurs when an object is partially visible, such as through foliage, exposing its surface at different shattered image regions. Even if the camera or object is moving, the object is never fully visible. 
In contrast to the human vision research community, we currently have very limited knowledge about vision under fragmented occlusion~\cite{black-CVIU1996, pflugfelder-avss2021, pegoraro-acvrw2020, darrell-WVM1991}. 
It turns out that humans can effortlessly perceive such occluded objects. Despite some advancements, we still do not comprehend this remarkable human ability, and current recognition algorithms do not provide an explanation either~\cite{zou-ieee2023}. Most algorithms are designed for the fully visible object. Even algorithms designed for recognition under partial occlusion~\cite{osherov-iccv2017, kortylewski-cvpr2020, zhang-ijcv2021, zhan-bmvc22, he-reports2024, kotwal-icip2024} would not work satisfactorily under fragmented occlusion, as they need sufficiently large visible regions to extract features.
Although fragmented occlusion seems like a niche problem, the study of it will enhance our overall understanding of occlusion~\cite{soatto-iclr2016}, which remains a challenge, still significantly impacting recent research~\cite{vishniakov-icml2024}.

As mentioned, psychologists have studied fragmented occlusion for a longer time. This area is known as dynamic perception~\cite{caplovitz-2021, palmer-2006, mccarthy-2017}.  They believe that our brain aggregates perceptual information over time and then reconstructs shape and appearance of such occluded objects. This paper is partially devoted to this belief.
The proposed 2D reconstruction algorithm takes a sequence of occluded images and provides a generated, integrated image where the object becomes more visible (see \cref{fig:reconstruction}). The motivation is that recognition algorithms are not applied to the single occluded images, but to this integrated image. Thus, in a sense, the paper provides both, a computational model for the human ability to perceive occluded objects and the reconstruction algorithm for making occluded objects more visible, which we call object de-occlusion.

Given the limited literature, a possible algorithm for object de-occlusion is to reconstruct the complete 3D scene with voxel coloring~\cite{seitz-cvpr1997} and then to delete the occluder voxels and project back all remaining scene elements into a reference camera~\cite{vaish-cvpr2006}. This approach is very sensitive to the unknown object's depth, which causes severe visible holes in the reconstruction.
Another way is to control the focal length and the aperture~\cite{favaro-cvpr2003}. Such an algorithm is effective and may suit a human computational model, but it is very much limited to short distances in the micrometer range~\cite{liu-cvpr2017}.
Several algorithms, denoted as the {\em synthetic aperture approach}, address the object de-occlusion by combining data from multiple cameras. Even for a moderate 70\,\% occlusion, many cameras are needed, where these are usually arranged in arrays~\cite{vaish-cvpr2004, vaish-cvpr2006} or lines~\cite{nathan-srs2022, neel-cvpr2007}; \eg 88 cameras are necessary in a foliage experiment~\cite{vaish-cvpr2006}.
Alternatively, a camera rig on a drone~\cite{kurmi-imaging2018} or camera on a computer controlled gantry~\cite{vaish-cvpr2006, wang-wacv2020, zhang-cvpr2021, yu-tpami2023} is used. Camera baselines are typically in the scale of the occluder size~\cite{kurmi-sensors2019}. Such apparatuses use RGB cameras~\cite{wang-wacv2020}, thermal, event cameras~\cite{zhang-cvpr2021, yu-tpami2023}, or combinations~\cite{liao-cvpr2022, li-mir2022} and apply neural networks for better reconstruction~\cite {wang-wacv2020, zhang-ijcai2021, hur-wacv2023}, but
the approach is generally limited by the need for static scenes and calibrated cameras, which are difficult to obtain in foliage.

This paper presents a further algorithm for object de-occlusion which uses video. This approach assumes a single uncalibrated camera and exploits the visual parallax between the camera and the object in the images of the sequence. The approach also assumes that the object parts become visible as they move. Our algorithm is conceptually simple: The images are first warped, so that the occluded object becomes stationary, and then the registered images are averaged together to create a single image containing the more visible object, while other image regions become blurred.
Despite occlusion, this is achieved using a parametric motion model~\cite{bergen-eccv1992} and by maximizing contrast within a {\em window in the integrated image} to robustly estimate the motion parameters.
We test this robustness on real video and synthetic images, using a new simulation that mimics foliage. We found that the algorithm reconstructs objects when they are barely visible, with an occlusion density of as high as 92.3\,\%.
We also combine this new algorithm for object de-occlusion with zero-shot object classification, resulting in a 68\,\% reduction in the classification error rate for such 90\,\% occluded images.

In summary, the contributions are as follows: 
\begin{enumerate}
\item An algorithm that combines global registration, averaging, and contrast maximization within a window in the  integrated image, demonstrating superior performance compared to the state-of-the-art for reconstructing occluded objects in both synthetic and real imagery
\item A fragmented occlusion simulation that is used for quantitatively evaluating the algorithm's behavior
\item Object classification, robust under fragmented occlusion, and its successful evaluation
\end{enumerate}

The paper is organized as follows: \cref{sec:relwork} discusses methods relevant for the proposed algorithm. \cref{sec:method} then describes the algorithm in detail. \cref{sec:experiments} presents the results of the experiments and discusses the findings, as well as describing the classification algorithm. \cref{sec:conclusions} concludes the paper with some analogies from psychological research.

\section{Related Work} \label{sec:relwork}
Registering sequentially acquired data and integrating it into a single representation has been previously discussed in a few papers: some studies of motion computation with two consecutive images~\cite{irani-ijcv1994} and with event data~\cite{gallego-cvpr2018, gallego-cvpr2019}, for the application of de-fencing~\cite{mu-tcsvt2014}, for the synthetic aperture approach with a thermal camera on a flying drone~\cite{kurmi-grsl2021}, and for the inverse synthetic aperture approach using radar~\cite{vehmas-access2021} or airborne thermal images~\cite{nathan-drones2022}.
Our algorithm differs from these papers by using a windowed contrast function to determine motion from the integrated image.

Joint 2D reconstruction and motion computation are already considered~\cite{kokaram-ivc2004, schoenemann-tip2012, liu-tpami2022, tsogkas-wacv2023}. These algorithms either alternate between optical flow estimation and reconstruction~\cite{schoenemann-tip2012, liu-tpami2022}, or solve both tasks simultaneously using a neural network~\cite{tsogkas-wacv2023}. 
The proposed algorithm, while potentially linked to these papers, focuses on fragmental occlusion. These earlier works fail to handle fragmental occlusion because they rely on optical flow, which is undefined for occluded pixels~\cite{ayvaci-ijcv2012}, or because of fundamental limitations of convolutional neural networks with occlusion~\cite{achille-phd2019}.

The work most similar to this paper is described in \cite{nathan-drones2022}. Their reconstruction relies on the infrared domain and fails with visible images.
Our approach differs from this paper in two ways: we use selective windowing on the occluded object to enhance contrast measurement, and we use a coherent motion model.
The proposed algorithm can now be applied to visible images and successfully de-occludes the reference image. In \cite{nathan-drones2022}, the authors manually set the motion parameters. We also introduce a new fragmental occlusion simulation to evaluate the algorithm. Finally, we demonstrate that classification under dense occlusion is feasible by feeding the integrated image into a standard classifier.

\section{Method} \label{sec:method}
Consider a scene of relative motion between a camera and an object. Let the image foreground contains all scene elements in front of the object, where some parts of the foreground occlude the object. As illustrated in \cref{fig:reconstruction}, the camera captures a sequence of images $I_1, \ldots, I_t, \ldots, I_T$, for which each image $I_t \in \mathbb{R}^{H \times W}$ is of height $H$ and width $W$. The stacked images $\mathcal{I} \in \mathbb{R}^{H \times W \times T}$ are the input to the algorithm. One of the images in the sequence is denoted the reference image $I_r$. The result of the algorithm is a computationally generated image $\hat{I}_r \in \mathbb{R}^{H \times W}$ which represents the reconstructed object at time $r$.

\subsection{Underlying Computation} 
\label{sec:theory}
\begin{description}[leftmargin=0pt]
\item[Motion Compensation:]
A classical constraint for computing the apparent motion between two images $I_t$ and $I_{t+1}$ is the image brightness constancy~\cite{horn-AI1981}
\begin{align} \label{eq:constancy1}
I_{t+1}(u + p(u, v, t), v + q(u, v, t)) \approx I_t(u, v).
\end{align}
Under most circumstances, the apparent brightness of moving objects remains constant~\cite{trucco-1998}.
The components $p(u, v, t)$ and $q(u, v, t)$ in \cref{eq:constancy1} form a so-called flow vector
\begin{align} \label{eq:flowvector}
\vec{v}_t(u, v) = \rowvec{p(u, v, t)}{q(u, v, t)}^\top
\end{align}
that describes the apparent motion in pixel $I_t(u, v)$. $\vec{v}_t(u, v)$ points to the corresponding location of this pixel in $I_{t+1}$. The flow vectors of all pixels specify a warping operation
\begin{align} \label{eq:flowfield}
\mathcal{F}_t(u, v) = \rowvec{u}{v}^\top + \vec{v}_t(u, v) 
\end{align}
between $I_t$ and $I_{t+1}$.
\cref{eq:constancy1} can be rewritten using a warping operation that takes $I_{t+1}$ and warps it according to $\mathcal{F}_t$ to get 
\begin{align} \label{eq:constancy2}
I_t(u, v) \approx I_{t+1}( \mathcal{F}_t(u, v)) = J_{t+1}(u, v).
\end{align}
$J_{t+1}$ in \cref{eq:constancy2} is a motion-compensated image with pixels registered to their corresponding pixels in $I_t$.

Without loss of generality and for simplicity, let $r = 1$. 
With this reference image, brightness constancy is applicable to the images $I_t ~ (1 < t \le T)$, by expanding \cref{eq:constancy2}:
\begin{align} \label{eq:constancy3}
I_r(u, v) = I_1(u,v) \approx J_t = I_t(\mathcal{F}_1 \circ \ldots \mathcal{F}_{t-1} (u, v)).
\end{align}
Let $J_1=I_1$: \cref{eq:constancy3} warps the $T-1$ images to the reference image. These motion-compensated images
and the reference image are then put together to form the image stack
\begin{align} \label{eq:stack}
\mathcal{J} = (J_1, J_2, \ldots, J_T) \in \mathbb{R}^{H \times W \times T}.
\end{align}
The advantage of the stack structure is that the registered pixels are slices  $(\mathcal{J}(u, v, 1), \ldots, \mathcal{J}(u, v, T))$ of $\mathcal{J}$, which makes the successive integration of images directly applicable. Without warping, the motion compensation would need to infer the trajectories from the flow fields beforehand and then retrieve the pixel values which is  memory inefficient.

\item[Image Integration:]
Following common techniques in various fields (\eg ~\cite{hassan-ejop2010},~\cite{irani-cvgip1991},~\cite{vaish-cvpr2006}), we propose to specify the integrated image as the average
\begin{align} \label{eq:int}
\hat{I}_r(u, v) =  \frac{1}{T} \sum_{t = 1}^T \mathcal{J}(u, v, t).
\end{align}

\cref{eq:constancy3} implies that brightness constancy is established between the reference image and the integrated image
\begin{align} \label{eq:constancy4}
I_r(u, v) \approx \hat{I}_r(u, v), \quad \rowvec{u}{v} \in \mathcal{O} \subseteq I_r.
\end{align}
Suppose the flow field corresponds to the relative motion between object and camera. Then, \cref{eq:constancy4} is valid for the correctly compensated motion and for region $\mathcal{O}$, which contains the object in the integrated image. Although averaging is sensitive to the occluder's brightness values, $\mathcal{O}$ does not appear blurry as much as pixels in the rest of the integrated image.
This is because the registered pixels outside $\mathcal{O}$ are averages of non-registered occluder and background. In contrast, the averages of pixels registered within $\mathcal{O}$ are closer to the unoccluded object appearance.

\item[Image Contrast:]
To find the warps $\mathcal{F}_t(u, v)$, we propose to evaluate the sample variance
\begin{align} \label{eq:var}
\mathrm{var}(\hat{I}_r) \triangleq \mathrm{var}(\{(\hat{I}_r)_i \}_1^{HW} )
\end{align}
over the $HW$ pixel values of the integrated image.

The pixel intensities in the integrated image results from warping and averaging and are not identical to intensities in the raw images.
For warping corresponding to the motion of some object,  the natural high contrast of this object is preserved also in $\hat{I}_r$. The empirical image variance is an indication of this contrast~\cite{gallego-cvpr2018}. This is especially true for objects of high contrast but seems to work in general.
The motion of a stationary occluder can be easily excluded because it is zero. If the sought-for occluded object is large enough, then the contrast is maximized for its nonzero motion, and enable the algorithm to estimate the required warping. 

On the other hand, the less accurate the warping is, the lower the contrast of the integrated image will be, because large areas in the integrated image are blurred due to the averaging process. This makes the sample variance smaller. 

This is a simple criterion but it has one important advantage: the contrast is a smooth function of the warping accuracy. This was confirmed in our experiments.  
Note that alternative, seemingly improved, criteria, such as the use of robust integration~\cite{vaish-cvpr2006} will lose this property. We also experimented with more complex gradient based cost functions (see  ~\cite{gallego-cvpr2019}) and found that they induce a harder search space, with many local maxima. 

\item[Fixation Window:]
We found that the sample variance for the correct warping may not be the maximum relative to other warpings. This is because real-world scenes often yield largely varying foreground and background intensities over the image. Consequently, the integrated sample variance is large and may overshadow the variance associated with the occluded object; see 
\crefrange{fig:park:zero}{fig:park:10} in \cref{sec:experiments} and \cref{sec:failurecase} in the supplemental material.

One way to mitigate this problem is to evaluate the sample variance over a smaller region or fixation window $W$ of the integrated image, which encompasses the object or object parts. That is to calculate the variance by \begin{align} \label{eq:var2}
\mathrm{var}_{|W|}(\{ \hat{I}_r(u, v) \}), (u, v) \in W \subseteq I_r, W \cap \mathcal{O} \neq \emptyset.
\end{align}
\cref{sec:evaluation} describes the procedure for computing this window.
\end{description}

\subsection{Algorithm}
\input{\figmethod{method}}

Using the variance to estimate the image contrast as an indication to the validity of the estimated flow fields, suggests an algorithm that alternates between computing $\hat{I}_r$ with \cref{eq:int} and (re-)estimating $\mathcal{F}_1, \ldots, \mathcal{F}_{t-1}$ by evaluating $\hat{I}_r$ with \cref{eq:var}. Thereby the reconstruction of the occluded object, which is within the integrated image, occurs simultaneously with the motion computation.
The (re-)estimation of the flow fields is written as optimization problem with the flow fields as unknown variables, or more compactly, as parameter vector
\begin{align}
\vectheta = \{ \vec{v}_1(1, 1)^\top, \vec{v}_1(2, 1)^\top, \ldots, \vec{v}_{T-1}(H, W)^\top \}. 
\end{align}
With $\vectheta$, \cref{eq:flowfield} can be written as parametrized warp
\begin{align} \label{eq:flowfield2}
\mathcal{F}_t(u, v; \theta) = \rowvec{u}{v}^\top + \rowvec{\vectheta_{k+1}}{\vectheta_{k+2}}^\top
\end{align}
with the vector's component index $k = 2HW(t-1) + H(v-1)+2(u-1)$. The optimization problem now is
\begin{align} \label{eq:learning}
\mathcal{\vectheta}^* &= \underset{\mathcal{\vectheta}}{\arg\max} ~ f(\vectheta) + \lambda R(\theta)
\end{align}
with objective function
\begin{align} \label{eq:objfun1}
f(\vectheta) = \mathrm{var}_{|W|}(\hat{I}_r).
\end{align}
Prior knowledge of the motion is encoded by the regularization term $R(\cdot)$. $\lambda$ is a hyperparameter.
By substituting $\hat{I}_r$ in \cref{eq:objfun1} with \cref{eq:int}, \cref{eq:stack} and \cref{eq:constancy3}, the objective function is written as
\begin{align} \label{eq:objfun2}
f(\vectheta) = \mathrm{var}_{|W|}(I_r + \sum_{t=2}^T I_t(\mathcal{F}_1 \circ \ldots \circ \mathcal{F}_{t-1} (u, v; \vectheta)).
\end{align}

If all warps $\mathcal{F}_t$ are accurate, \ie the result of the optimization $\vectheta^*$ is close to the true object motion, the objective function in \cref{eq:objfun2} will attain its maximum value. As explained above, the correct motion estimate  results in high contrast of the integrated image or a large value of $f(\theta)$. Conversely, all other motions cause blur, leading to low contrast.
\cref{fig:method} illustrates the algorithm: $\vectheta^*$ is found at $\nabla f = {\bf 0}$
and negative definite Hessian $H(f)$.

\subsection{Remarks} \label{sec:evaluation}
\begin{description}[leftmargin=0pt]

\item[Parametrized Motion:]
A completely free local motion model $\vec{v}_t(u, v)$ in \cref{eq:flowvector} would require too many variables. 
This work therefore uses a parametrized motion model~\cite{irani-ijcv1994}
\begin{align} \label{eq:flowvector2}
\vec{v}_t(u, v; \vectheta) = \rowvec{p(u, v, t; \vectheta)}{q(u, v, t; \vectheta)}^\top.
\end{align}
The flow components $p$ and $q$ describe the motion in \cref{eq:flowvector2} by some functions, typically low dimensional polynomials. Certainly, $p$ and $q$ must be complex enough to capture real motion. A second advantage of the parametrized motion models over the local model is that their computation is more robust to fragmented occlusion. 

In this paper we use a motion model of coherent translation at certain depth, which is a special case of a rigid scene's translation~\cite{trucco-1998}:
\begin{align}
p(u, v, t; \vectheta) & = (t-r) \alpha \label{eq:transp}\\
q(u, v, t; \vectheta) & = (t-r) \beta \label{eq:transq}
\end{align}
with two parameters
\begin{align} \label{eq:theta}
\vectheta = \rowvec{\alpha}{\beta}^\top. 
\end{align}
The translation in the scenes considered here is often close to horizontal in the image plane, which encourages the prior
\begin{align}
R(\vectheta) = -\beta^2.
\end{align}
This simpler motion model seems sufficient for the general assumption of an upright camera and objects moving at close to constant depth on a perpendicular ground plane, and using it led to successful reconstructions with videos acquired by a camera.

\item[Implementation Details:]
We used two alternative method for optimization. For visualization of $f$, we use grid search over intervals of $\vectheta$. For more efficient search, we parallelize the problem in JAX~\cite{frostig-mlsys2018}. We use gradient ascent~\cite{baydin-JMLR2018} with Adam in combination with JAX's accurate auto-differentiation. For the experiments, we set $\lambda$ to $\frac{1}{2}$ and the step size to $0.1$. 

\item[Window Placement:]
As discussed above, for highly varying brightness in outdoor images, it is crucial to estimate the variance within a window and not over the whole image. The fixation window is placed at the location of maximal saliency, and its size is set to 10\,\% of the image size. The work uses a simple motion based saliency map which is the absolute difference of consecutive images. For good results, the images are blurred with a Gaussian kernel. 

\end{description}

\section{Experimental Results} \label{sec:experiments}
The experimental methodology follows Vaish~\etal's empirical analysis~\cite{vaish-cvpr2006} and includes experiments with synthetic images and images acquired by a camera.
Selected videos are provided in the supplementary material (see \cref{sec:videos}).

\subsection{Reconstruction in Synthetic Videos}
Following the scene model outlined in \cref{sec:method}, we generate eight-bit, grayscale images using Luxor (\url{https://juliapackages.com/p/luxor}). \cref{fig:reconstruction} is an example. The background behind the object and the foreground, which also comprises the occluder, are set to a mid-gray value of $128$. The foreground is composed of randomly placed and oriented triangles that may overlap one another~\cite{bridson-siggraph2007}. Uniformly varying the triangles' size controls the occlusion density $\rho$ (\%) up to a small, random deviation. This occlusion model mimics foliage reasonably. For a different model see~\cite{kurmi-sensors2019}. 

\begin{description}[leftmargin=0cm]
\item[Experimental Setup:]
As the object, we choose a planar "Ape" cartoon from OpenClipArt (\url{https://openclipart.org}), convert it to grayscale and resize the image to fill two-thirds of the $256^2$\,px canvas. For each generated image sequence, the object appears at the image center for the reference time
\begin{equation}
r = \frac{1}{2}(T-1), \quad T = 2t+1, \quad t = 1, 2, 3, \ldots,
\end{equation}
and the object is following a uniform translation in the image plane. We randomly varied velocity within fixed bounds to generate image sequences of different length $T$.
The experimental methodology, but not the algorithm, had access to the unoccluded image $\bar{I}_r$, to a binary mask $M_r$ indicating all the object's pixels and to a measurement of total visibility $\nu$ (\%) that expresses the fraction of pixels that are visible in at least one image of the sequence. 

We generate a total of 300 sequences for six specified occlusion levels: $\rho = \{10, 30, 50, 70, 80, 90\}$.  Each sequence contains a unique occlusion pattern, specified by the triangles' size, positions and orientations. Consequently, the actual occlusion level is random, and given in \cref{tab:statistics}.

\begin{table}
\centering
\begin{tabular}{c|c|c|c}
 & $\rho$ & $T$ & $\nu$ \\
\hline
1 & $11.67 \pm  0.50$ & $25.24 \pm 12.32$ & $100.0 \pm 0.00$ \\
\hline
2 & $28.91 \pm 1.01$ & $25.40 \pm 12.26$ & $100.0 \pm 0.00$ \\
\hline
3 & $51.20 \pm 1.11$ & $23.36 \pm 09.12$ & $100.0 \pm 0.00$ \\
\hline
4 & $70.22 \pm 1.59$ & $24.44 \pm 10.95$ & $99.83 \pm 2.42$ \\
\hline
5 & $82.57 \pm 1.14$ & $23.98 \pm 10.66$ & $96.64 \pm 2.61$ \\
\hline
6 & $90.51 \pm 1.87$ & $25.91 \pm 12.13$ & $84.00 \pm 8.69$ \\
\end{tabular}
\caption{Synthetic video statistics: The image sequences are divided into six sets associated, each, with roughly the same occlusion density ~$\rho$ ($\text{mean} \pm \text{std. deviation}$ of each quantity). The sequence length $T$ and the object's total visibility $\nu$ are also given.}
\label{tab:statistics}
\end{table}

\item[Metrics:] To assess the quality of the reconstructed image, we calculate the Pearson correlation coefficient
\begin{align} \label{eq:pearson1}
\mu_{\hat{\text{P}}} \doteq \mu_\text{P}(M_r\hat{I}_r, M_r\bar{I}_r)
\end{align}
between the reconstructed image $\hat{I}_r$ and the known, unoccluded image $\bar{I}_r$. As the algorithm reconstructs the object and no other scene element, the metric is only applied to the object region, determined by the mask $M_r$.

For comparison, we include the correlation coefficient
\begin{align} \label{eq:pearson2}
\mu_{\text{P}} \doteq \mu_\text{P}(M_r I_r, M_r\bar{I}_r)
\end{align}
between the given occluded image ${I}_r$ and $\bar{I}_r$.
 
We also use the  mean-squared error (MSE) ratio as an additional measure of quality for the reconstruction: 
 \begin{equation}
\mu_\text{MSE}  = \frac{\text{MSE}(M_rI_r, M_r\bar{I}_r)- \text{MSE}(M_r\hat{I}_r, M_r\bar{I}_r)}{\text{MSE}(M_rI_r, M_r\bar{I}_r)}.
\end{equation}

$\mu_\text{P}$ and $\text{MSE}$ are described in \cref{sec:measures} of the supp. material.

\item[Results:]
\cref{fig:ape50} shows the experimental results. 
Note that the average correlation coefficient associated with the reconstructed images $\mu_{\hat{\text{P}}}$ is larger than $0.7$ for all occlusion levels. On the other hand, as expected, the correlation coefficient associated with the original occluded images  $\mu_\text{P}$ is lower than $0.23$. The large values of $\mu_{\hat{\text{P}}}$, obtained after the integration, is a strong indicator that corresponding pixels are found in the warped images of \cref{eq:stack} - even for very dense occlusion. The MSE ratio rapidly decreases as the level of occlusion increases. 
Some of the degradation in reconstruction performance is due to object parts that are never seen during the sequence; see $\nu$ in \cref{tab:statistics} for sets 4-6. 

\begin{figure}
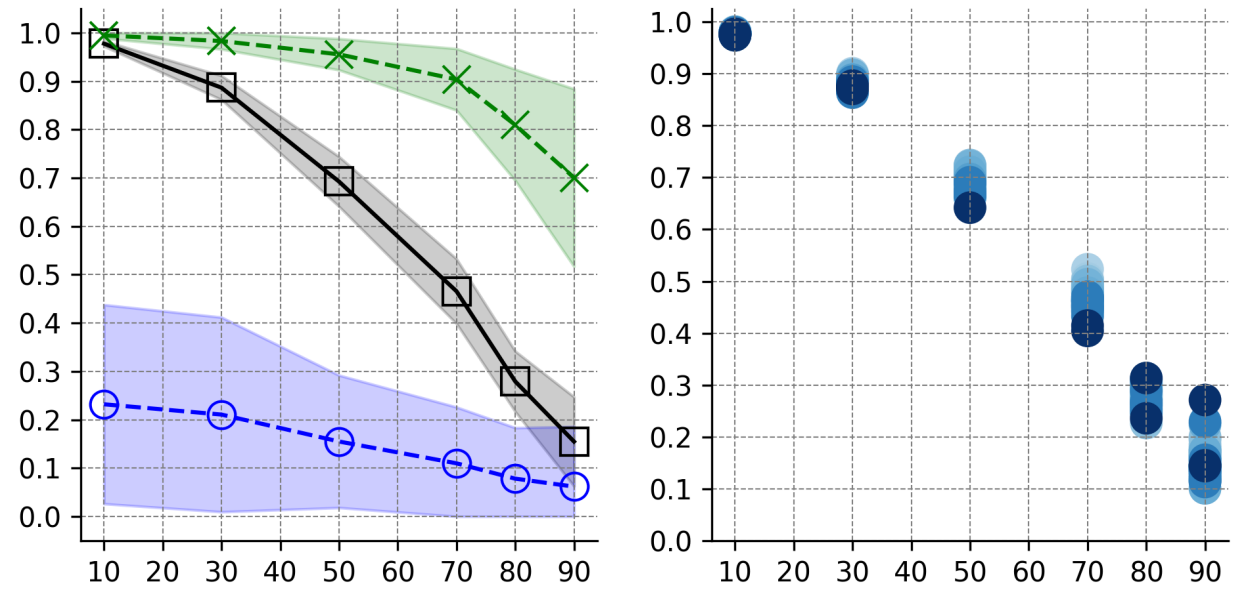

\centering
\begin{subfigure}[b]{.49\linewidth}
\includegraphics[width=\textwidth]{\figexperiment{synex-avg-trans}}
\caption{Results}
\end{subfigure}
\begin{subfigure}[b]{.49\linewidth}
\includegraphics[width=\textwidth]{\figexperiment{synex-avg-trans-seqlen}}
\caption{Evaluation of $\mu_\text{MSE}$ at $T$}
\end{subfigure}
\caption{Experimental results: (a) The plots draw $\mu_\text{MSE}$ (square), $\mu_{\hat{\text{P}}}$ (\textcolor{ForestGreen}{cross}), $\mu_{\text{P}}$ (\textcolor{blue}{circle}) with confidence interval $\pm 3\sigma$ (ordinate)  \vs $\rho$ (abscissa). (b) Again $\mu_\text{MSE}$ \vs $\rho$, but we provide information on the sequence length $T$: Dark blue encodes long sequences. Longer sequences do not necessarily leader to a better reconstruction.}
\label{fig:ape50}
\end{figure}

\crefrange{fig:ape50:057}{fig:ape50:300} illustrates occluded and reconstructed images for five representative samples of the 300 synthetic videos. The quality of the reconstructed object is comparable to the result in~\cite{vaish-cvpr2006}. Their plane sweeping algorithm, however, fails for $\rho > 75\,\%$, while the motion computation in \cref{eq:learning} is successful even for $\rho > 90\,\%$.
This success is also evident from the large $\mu_{\hat{\text{P}}}$ values. The gradient-based motion estimation algorithm in \cref{eq:learning} achieves an accuracy of $\alpha, \beta \pm 0.01\,\text{px}$ or better. \cref{fig:ape50:results} lists the associated quantitative characteristics of these five sample sequences. 

\begin{figure}
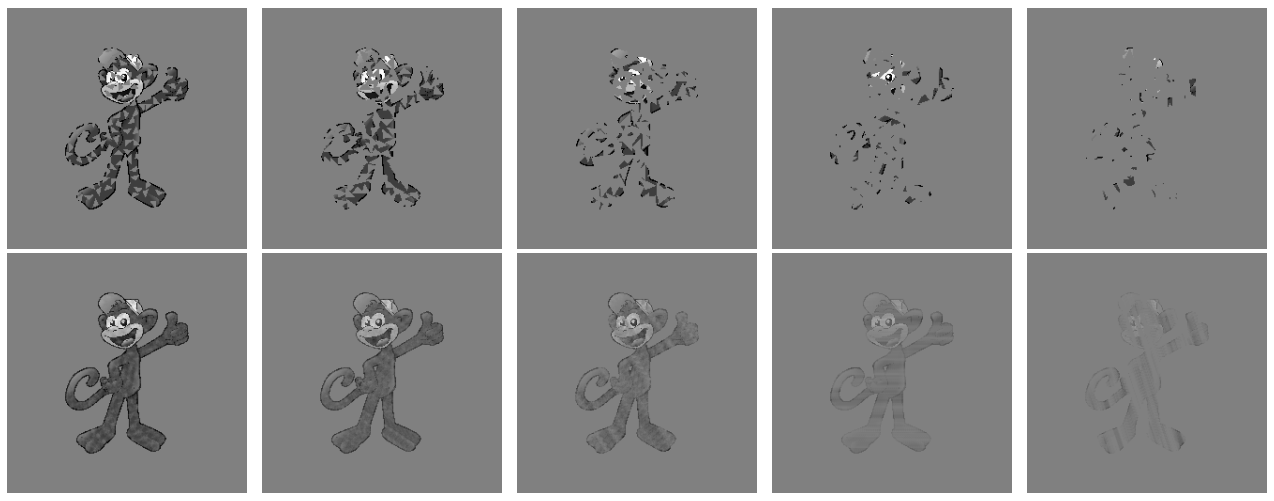

\centering
\begin{subfigure}[b]{\linewidth}
\centering
\footnotesize
\begin{tabular}{r|r|r|r||r|r||r|r|r}
\multicolumn{1}{c}{\small \#} & \multicolumn{1}{c}{\small $\rho$} & \multicolumn{1}{c}{\small $T$}  & \multicolumn{1}{c}{\small $\nu$} & \multicolumn{1}{c}{\small $\alpha$} & \multicolumn{1}{c}{\small $\beta$} & \multicolumn{1}{c}{\small $\mu_\text{MSE}$} & \multicolumn{1}{c}{\small $\mu_{\hat{\text{P}}}$} & \multicolumn{1}{c}{\small $\mu_{\text{P}}$}\\
\hline
$57$ & $27.6$ & $13$ & 100 & -6.0 & -6.0 & $0.90$ & $0.97$ & $0.09$\\ 
149 & 49.7 & 19 & 100 & 4.0 & 6.0 &  0.72 & 0.97 & 0.10\\
193 & 70.3 & 13 & 99 & -6.0 & -4.0 & 0.52 & 0.87 & 0.02\\
203 & 85.2 & 59 & 99.8 & 0.0 & 2.0 & 0.27 & 0.86 & 0.16\\
300 & 92.3 & 25 & 71.2 & -3.0 & -1.0 & 0.10 & 0.54 & 0.07\\
\end{tabular}
\caption{Statistics and quantitative results}
\label{fig:ape50:results}
\end{subfigure}\\[1ex]
\begin{subfigure}[b]{.19\linewidth}
\includegraphics[width=\textwidth]{\figexperiment{057-synex-s_3_0_-n_07_-0000000007}}
\includegraphics[width=\textwidth]{\figexperiment{016-057-synex-s_3_0_-n_07_-msi-u_-06_00_-v_-06_00_}}
\caption{27.6\,\%}
\label{fig:ape50:057}
\end{subfigure}
\hfill
\begin{subfigure}[b]{.19\linewidth}
\includegraphics[width=\textwidth]{\figexperiment{149-synex-s_5_0_-n_49_-0000000010}}
\includegraphics[width=\textwidth]{\figexperiment{178-149-synex-s_5_0_-n_49_-msi-u_+04_00_-v_+06_00_}}
\caption{49.7\,\%}
\label{fig:ape50:149}
\end{subfigure}
\hfill
\begin{subfigure}[b]{.19\linewidth}
\includegraphics[width=\textwidth]{\figexperiment{193-synex-s_7_0_-n_43_-0000000007}}
\includegraphics[width=\textwidth]{\figexperiment{018-193-synex-s_7_0_-n_43_-msi-u_-06_00_-v_-04_00_}}
\caption{70.3\,\%}
\label{fig:ape50:193}
\end{subfigure}
\hfill
\begin{subfigure}[b]{.19\linewidth}
\includegraphics[width=\textwidth]{\figexperiment{203-synex-s_9_0_-n_03_-0000000030}}
\includegraphics[width=\textwidth]{\figexperiment{114-203-synex-s_9_0_-n_03_-msi-u_+00_00_-v_+02_00_}}
\caption{85.2\,\%}
\label{fig:ape50:203}
\end{subfigure}
\hfill
\begin{subfigure}[b]{.19\linewidth}
\includegraphics[width=\textwidth]{\figexperiment{300-synex-s_11_0_-n_50_-0000000013}}
\includegraphics[width=\textwidth]{\figexperiment{066-300-synex-s_11_0_-n_50_-msi-u_-03_00_-v_-01_00_}}
\caption{92.3\%}
\label{fig:ape50:300}
\end{subfigure}
\caption{Five synthetic sample videos: (a) quantitative data. (b) - (f) Raw occluded images (top row) and the object reconstructions for the different occlusions densities (bottom row).}
\label{fig:ape}
\end{figure}
7
\subsection{Robust Recognition} \label{sec:recognition}
Visually, it appears that the images are more comprehensible after reconstruction (see \crefrange{fig:ape50:057}{fig:ape50:300}). To quantify this impression, the following experiment shows that the proposed algorithm enhances the resilience of zero-shot off-the-shelf classification to fragmented occlusion.

\begin{description}[leftmargin=0cm]
\item[Experimental Setup:] We use the CLIP model~\cite{radford-pmlr2021} based on the pretrained ViT-B/32 architecture (\url{https://huggingface.co/openai/clip-vit-base-patch32}) to encode both occluded and reconstructed images, one from each of the 300 sequences.  We then specify text labels, such as: \texttt{"a photo of a \{$l$\}"}, with labels $l \in L$ coming from a set $L$ of category names, including \texttt{ape} and diverse outdoor categories such as \texttt{deer}, \texttt{helicopter}, \texttt{swamp}. These categories are inspired by~\cite{connor-eccv2024} and adapted to our project needs. The 24 used text labels are found in \cref{sec:labels} of the supp. material. Finally, we encode the tokenized text labels into the same CLIP space. No training is done.

\item[Metric:] Let $\vec{p}$ and $\vec{q}$ be two features of image and text respectively. 
The cosine distance~\cite{salton-acm1975} between them is
\begin{align} \label{eq:cosinedist}
d(\vec{p}, \vec{q}) = 1 - \frac{\vec{p} \vec{q}^\top}{\|\vec{p} \| \|\vec{q}\|}. 
\end{align}
Due to CLIP's modality gap~\cite{liang-neurips2022}, the cosine distance is typically large. The following normalized measure is context-aware and scale-invariant, and better accounts for the contrastive loss employed during CLIP training~\cite{radford-pmlr2021}
\begin{align}
D(\vec{p}, \vec{q}_r) = \frac{1}{ |L| } \sum_i \frac{d(\vec{p}, \vec{q}_r)}{d(\vec{p}, \vec{q}_i)}.
\end{align}
$D$ determines whether a text label $\vec{q}_r$, when compared to all other text labels, is relatively distant or close to image $\vec{p}$. 

\item[Results:] \crefrange{fig:rec:occluded}{fig:rec:reconstructed} evaluate $D$ for all $300$ occluded and integrated images, and show the means and the $2\sigma$
confidence intervals of the correct label \texttt{ape}.
The results confirm our first impression: The proposed algorithm substantially reduces the normalized distance $D$ for the \texttt{ape} label. 

Top1 classification with the occluded image 
fails for $\rho > 30\,\%$, 
but, for deoccluded images,  was perfect for $\rho < 90\,\%$ (see \cref{fig:rec:errorrate}). 
The same results were (obviously) obtained with the cosine distance. 
This is useful and remarkable, when remembering that usually, CNN models trained on non-occluded images, fail under occlusion \cite{osherov-iccv2017}. 

\end{description}

\begin{figure}
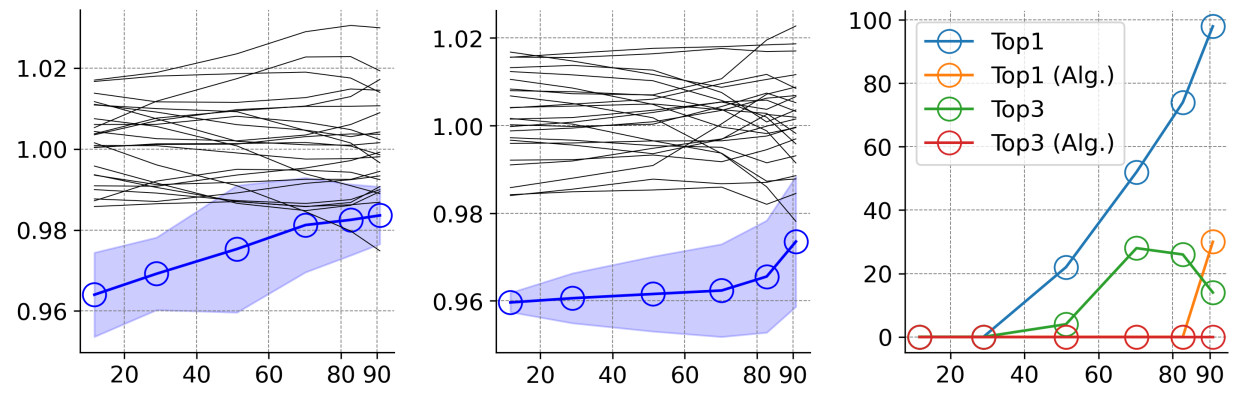

\begin{subfigure}[b]{.32\linewidth}
\includegraphics[width=\textwidth]{\figexperiment{clip-cosine-sim-occluded}}
\caption{Occluded}
\label{fig:rec:occluded}
\end{subfigure}
\begin{subfigure}[b]{.32\linewidth}
\includegraphics[width=\textwidth]{\figexperiment{clip-cosine-sim-reconstructed}}
\caption{Reconstructed}
\label{fig:rec:reconstructed}
\end{subfigure}
\begin{subfigure}[b]{.32\linewidth}
\includegraphics[width=\textwidth]{\figexperiment{clip-classification-error-rate}}
\caption{Cls. Error Rate}
\label{fig:rec:errorrate}
\end{subfigure}
\caption{Recognition with CLIP: (a) The distance $D$ \vs the occlusion density $\rho$ for the occluded images. The mean and confidence interval of \texttt{ape} is shown in blue. (b) The same evaluation for the reconstructed images, shows a significant improvement. (c) The CLIP's classification error rate \vs $\rho$.}
\label{fig:rec}
\end{figure}

\end{description}

\subsection{Foliage Experiments}\label{sec:foliage}
Three experiments demonstrate the proposed reconstruction in challenging real-world scenes. The first sequence contains 20 images, with a walking person in a park on a sunny day. 
The foliage is moderate and comparable to Vaish's foliage experiment~\cite{vaish-cvpr2006}. A second sequence contains 20 images of a closer person in the boreal forest with higher occlusion density. A third boreal forest sequence contains  30 images and was captured by panning the camera while the person was walking.

\begin{figure}
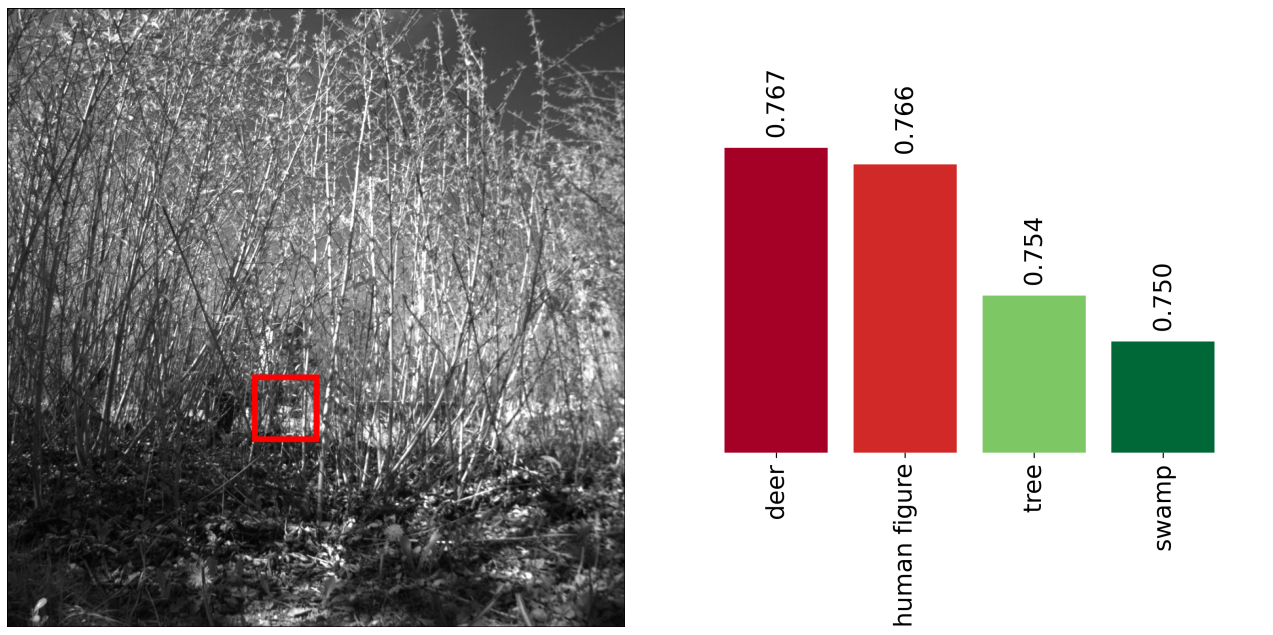
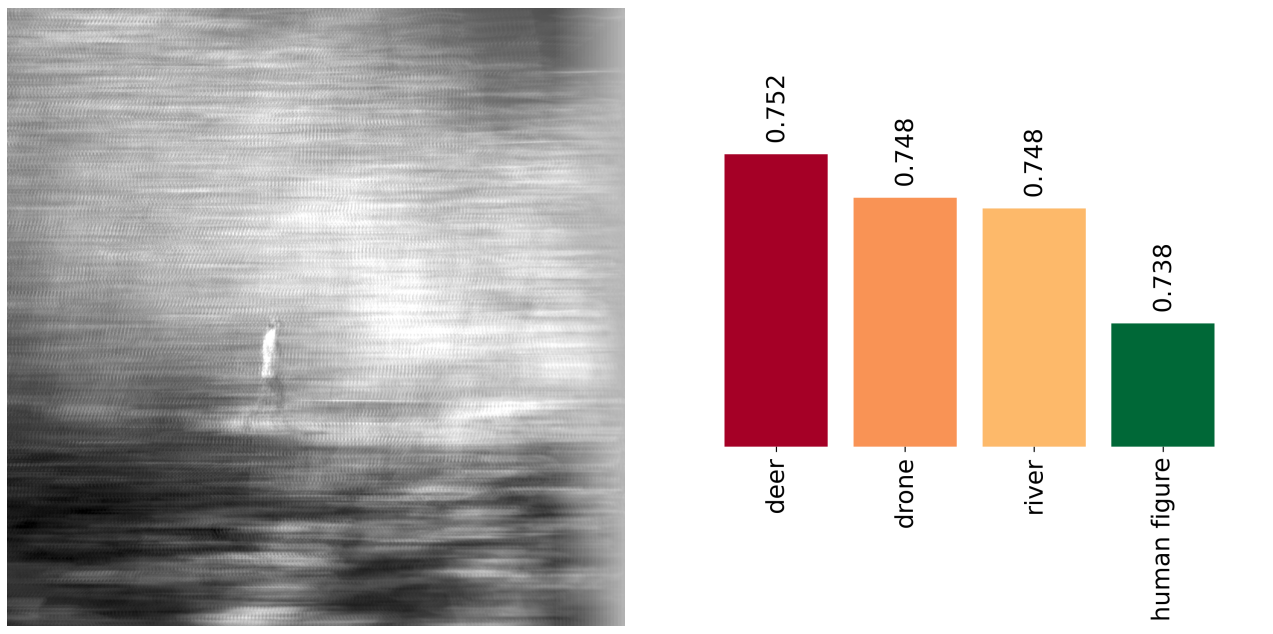
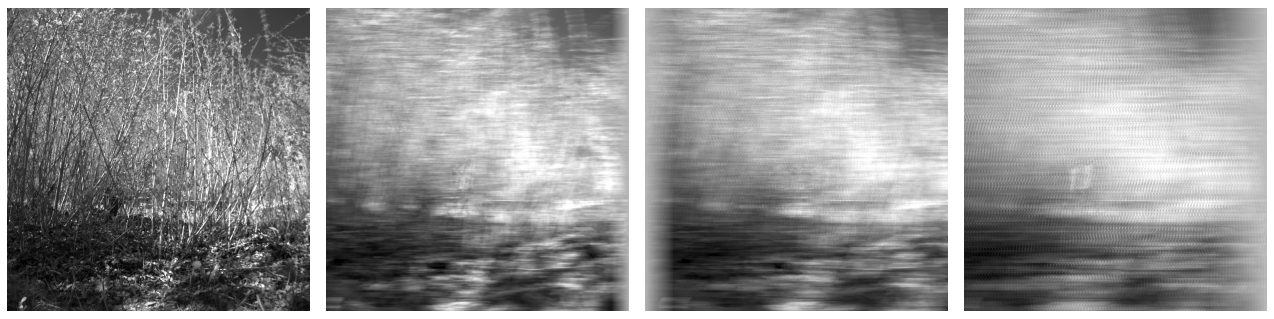

\centering
\begin{subfigure}[b]{.49\linewidth}
\includegraphics[width=\textwidth]{\figexperiment{tum-ref}}
\caption{$I_1$}
\label{fig:park:window}
\end{subfigure}
\hfill
\begin{subfigure}[b]{.45\linewidth}
\includegraphics[width=0.95\textwidth]{\figexperiment{clip-classification-foliage}}
\caption{Evaluation of $d$ with $I_r = I_1$}
\label{fig:park:occluded}
\end{subfigure}\\[1ex]
\begin{subfigure}[b]{.49\linewidth}
\includegraphics[width=\textwidth]{\figexperiment{064-tum-msi-v_+06_00_}}
\caption{$\alpha = 6.0\,\text{px}$}
\label{fig:park:person}
\end{subfigure}
\hfill
\begin{subfigure}[b]{.45\linewidth}
\includegraphics[width=.95\textwidth]{\figexperiment{clip-alg-classification-foliage}}
\caption{Evaluation of $d$ with $\hat{I}_r$}
\label{fig:park:reconstructed}
\end{subfigure}\\[1ex]
\begin{subfigure}[b]{.24\linewidth}
\includegraphics[width=\textwidth]{\figexperiment{040-tum-msi-v_+00_00_}}
\caption{$\alpha = 0.0\,\text{px}$}
\label{fig:park:zero}
\end{subfigure}
\hfill
\begin{subfigure}[b]{.24\linewidth}
\includegraphics[width=\textwidth]{\figexperiment{052-tum-msi-v_+03_00_}}
\caption{$\alpha = 3.0\,\text{px}$}
\label{fig:park:3}
\end{subfigure}
\hfill
\begin{subfigure}[b]{.24\linewidth}
\includegraphics[width=\textwidth]{\figexperiment{020-tum-msi-v_-05_00_}}
\caption{$\alpha = -5.0\,\text{px}$}
\end{subfigure}
\hfill
\begin{subfigure}[b]{.24\linewidth}
\includegraphics[width=\textwidth]{\figexperiment{080-tum-msi-v_+10_00_}}
\caption{$\alpha = 10.0\,\text{px}$}
\label{fig:park:10}
\end{subfigure}
\caption{Park results: (a) Reference image and fixation window. (b) Applying CLIP to (a). The bar plot shows the four smallest cosine distances for label \texttt{human figure} and the same 24 diverse outdoor categories as used for the synthetic videos. (c) object reconstruction, (d) applying CLIP to (c). Note the correct ranking compared to (b). (e) fore- and background reconstruction. (f) - (h) integral images corresponding to some incorrect motions.}
\label{fig:park}
\end{figure}

\begin{figure}
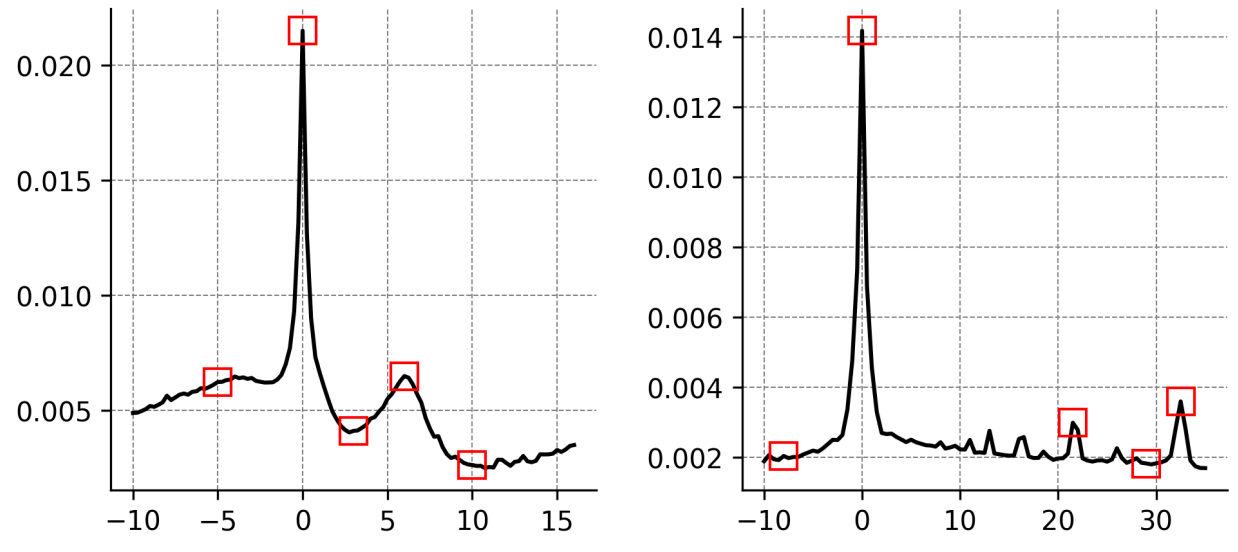

\centering
\begin{subfigure}[b]{.49\linewidth}
\includegraphics[width=\textwidth]{\figexperiment{tum-function}}
\caption{Park}
\label{fig:func:park}
\end{subfigure}
\begin{subfigure}[b]{.49\linewidth}
\includegraphics[width=\textwidth]{\figexperiment{sfence-grey-function}}
\caption{Fence}
\label{fig:func:fence}
\end{subfigure}
\caption{Both plots draw $f$ (ordinate) \vs $\alpha$ (abscissa). (a) Two significant maxima of the person and static motions are observable. (b) The overlaps of fence slats cause misleading maxima.}
\label{fig:func}
\end{figure}

\begin{figure}
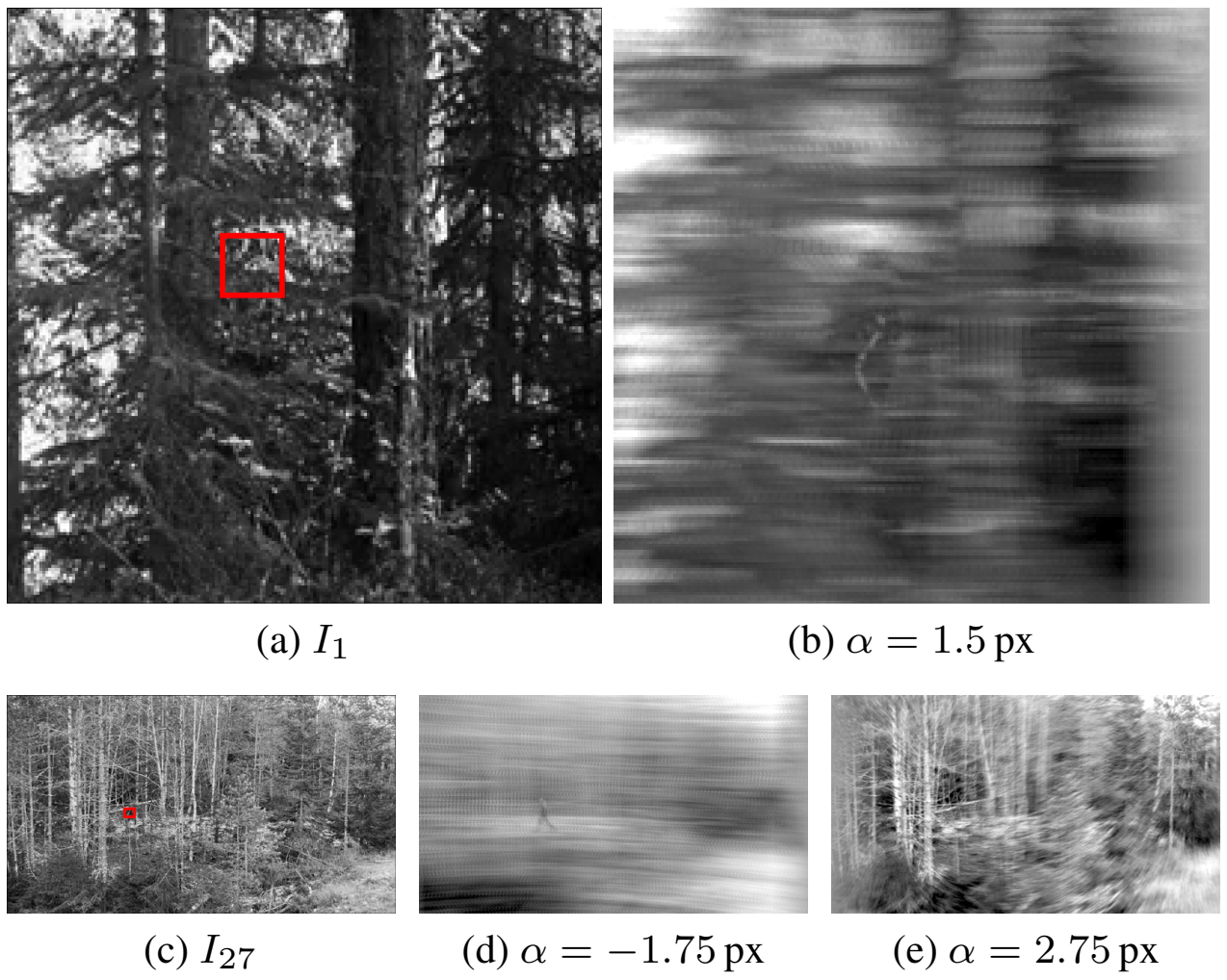

\begin{subfigure}[b]{.49\linewidth}
\includegraphics[width=\textwidth]{\figexperiment{fin-ref}}
\caption{$I_1$}\label{fig:boreal:window}
\end{subfigure}
\begin{subfigure}[b]{.49\linewidth}
\includegraphics[width=\textwidth]{\figexperiment{025-fin-msi-v_+01_50_}}
\caption{$\alpha = 1.5\,\text{px}$}\label{fig:boreal:person}
\end{subfigure}\\[1ex]
\begin{subfigure}[b]{.32\linewidth}
\includegraphics[width=\textwidth]{\figexperiment{kongsberg-ref}}
\caption{$I_{27}$}\label{fig:pan:window}
\end{subfigure}
\hfill
\begin{subfigure}[b]{.32\linewidth}
\includegraphics[width=\textwidth]{\figexperiment{005-kongsberg-msi-v_-01_75_}}
\caption{$\alpha = -1.75\,\text{px}$}\label{fig:pan:person}
\end{subfigure}
\hfill
\begin{subfigure}[b]{.32\linewidth}
\includegraphics[width=\textwidth]{\figexperiment{023-kongsberg-msi-v_+02_75_}}
\caption{$\alpha = 2.75\,\text{px}$}\label{fig:pan:trees}
\end{subfigure}
\caption{Boreal forest and camera pan results: (a) and (c) ref. images and fixation windows. (b) and (d) object reconstruction. (e) foreground (tree) reconstruction related to camera panning.}
\end{figure}

\begin{description}[leftmargin=0cm]
\item[Results:]
By employing the fixation window (\cref{fig:park:window,fig:boreal:window,fig:pan:window}), the algorithm finds a maximum for the person in all three sequences. Our current window placement algorithm, is simple and assumes a static camera. For the third sequence with camera motion, we therefore placed the fixation window manually.

\cref{fig:func:park} depicts an evaluation of $f$, clearly demonstrating in the Park sequence the maximum value for $\alpha = 6.0\,\text{px}$. The associated integrated image is in \cref{fig:park:person} and its histogram in \cref{fig:histopark} of the supp. material. Note that a peak for zero motion ($\alpha = 0$) in $f$ always exists, reflecting the static camera and the stationary large background and foreground; see the reconstructed static fore- and background integrated image in \cref{fig:park:zero}. 
To search for the correct person's motion, the optimizer in \cref{eq:learning} therefore excludes the subspace in the vicinity of $\alpha = 0$.

We also applied the CLIP classification algorithm to the park sequence. As with the synthetic videos, the same 24 categories were considered, plus \texttt{human figure} instead of \texttt{ape}. Using the de-occluded image as input led to  successful top-1 classification (\cref{fig:park:occluded,fig:park:reconstructed}).
\crefrange{fig:park:zero}{fig:park:10} illustrate qualitative results for specific translations. \cref{fig:park:person,fig:boreal:person,fig:pan:person} show the reconstructed persons.
To enhance visibility, those images’ histograms were linearly stretched to span the full brightness range 
$[0, 255]$. 

In contrast, in the third boreal sequence, the camera pan introduces a global motion. If this motion is selected then the algorithm reconstructs near distant trees which move at $\alpha = 2.75\,\text{px}$ (\cref{fig:pan:trees}).  The camera pan is not a translation, so not all parts of the foreground and background are sharply reconstructed.
\end{description}

\begin{figure}
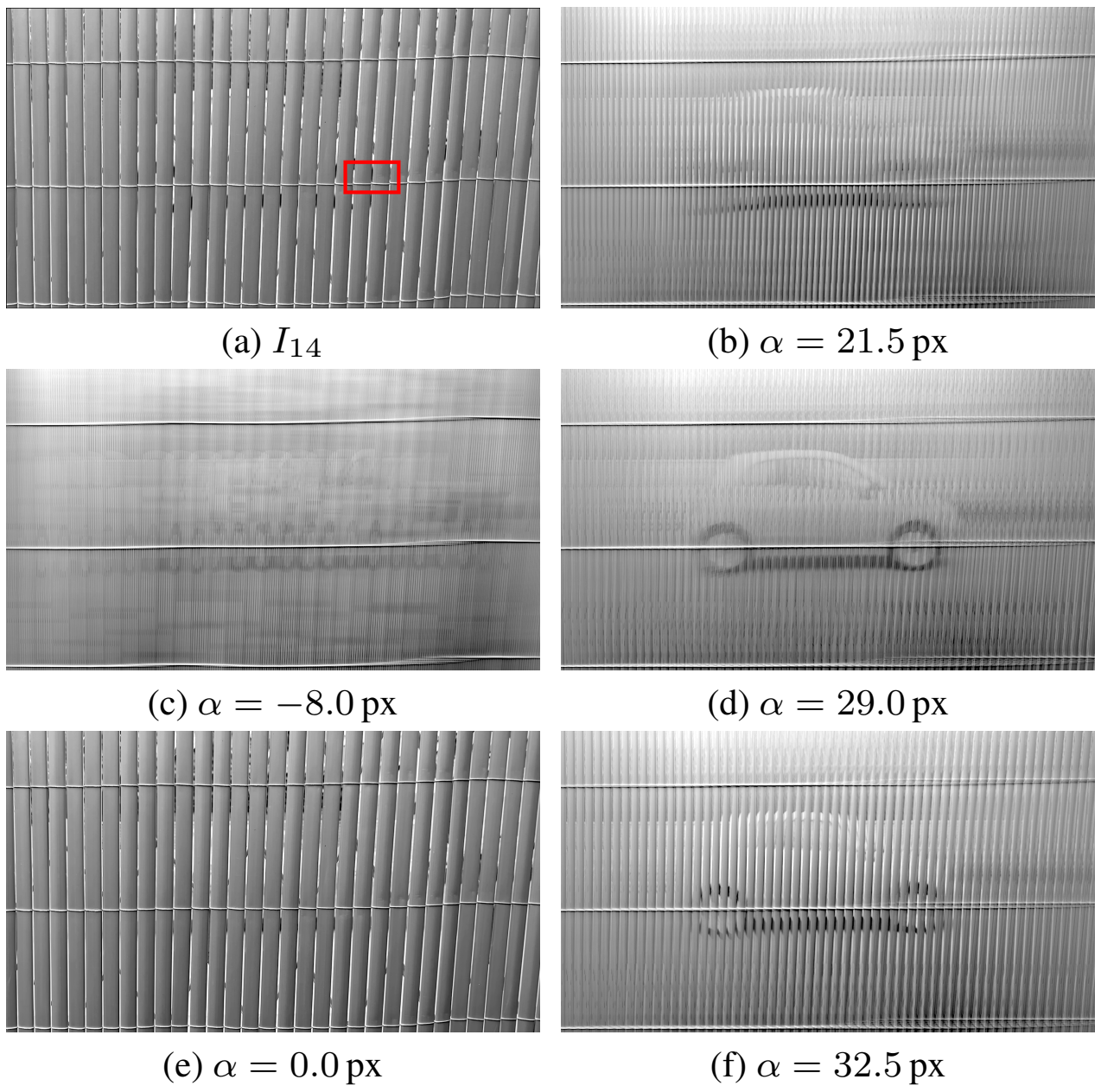

\begin{subfigure}[b]{.49\linewidth}
\includegraphics[width=\textwidth]{\figexperiment{sfence-grey-ref}}
\caption{$I_{14}$}
\end{subfigure}
\hfill
\begin{subfigure}[b]{.49\linewidth}
\includegraphics[width=\textwidth]{\figexperiment{063-sfence-grey-msi-v_+21_50_}}
\caption{$\alpha = 21.5\,\text{px}$}
\end{subfigure}
\begin{subfigure}[b]{.49\linewidth}
\includegraphics[width=\textwidth]{\figexperiment{002-sfence-grey-msi-v_-08_00_}}
\caption{$\alpha = -8.0\,\text{px}$}
\end{subfigure}
\hfill
\begin{subfigure}[b]{.49\linewidth}
\includegraphics[width=\textwidth]{\figexperiment{078-sfence-grey-msi-v_+29_00_}}
\caption{$\alpha = 29.0\,\text{px}$}
\label{fig:fence:29}
\end{subfigure}
\begin{subfigure}[b]{.49\linewidth}
\includegraphics[width=\textwidth]{\figexperiment{020-sfence-grey-msi-v_+00_00_}}
\caption{$\alpha = 0.0\,\text{px}$}
\end{subfigure}
\hfill
\begin{subfigure}[b]{.49\linewidth}
\includegraphics[width=\textwidth]{\figexperiment{085-sfence-grey-msi-v_+32_50_}}
\caption{$\alpha = 32.5\,\text{px}$}
\label{fig:fence:car}
\end{subfigure}
\caption{Fence results: (a) Reference image and fixation window. (b) - (f) Integral images for different motions. (e) fore- and background reconstruction. (f) car reconstruction.}
\label{fig:fence}
\end{figure}

\subsection{Fence Experiment}\label{sec:fence}
We finally tested the algorithm on a forth sequence of 30 images, with a moving car (Fiat 500) behind a garden fence. \cref{fig:fence} shows the results. This sequence was taken with a hand-held cellular phone, demonstrating the robustness to small camera motions during image acquisition.

As depicted in \cref{fig:func:fence}, $f$ exhibits numerous local maxima that are not associated to the car’s motion. 
Although the integrated image at $\alpha = 29$ has lower contrast (see \cref{fig:fence:29}), it appears to be a better reconstruction than the image found by the algorithm (see \cref{fig:fence:car}). 
The misleading maxima are probably due to the repetitive textures and regions of uniform intensities, such as the fence slats. See also the associated histograms in \cref{fig:histofence} and the of the supp. material.

\subsection{Discussion of Experimental Results}
The reconstruction quality is comparable to that obtained from synthetic apertures using stereo~\cite{vaish-cvpr2006}, but our algorithm maintains its accuracy for higher occlusion densities, exceeding $70\,\%$.
This is due to the proposed coherent motion computation, which is more robust to occlusion. We found that the quality primarily depends on the scene specific motion and occlusion patterns, rather than the video length, which is known for a long time~\cite{irani-ijcv1994}.

A classification algorithm, combining the proposed reconstruction and a zero-shot CLIP classification, achieves a high, 70.0\,\% top1 accuracy for images with approximately 90\,\% occlusion.
This is remarkable because applying CLIP on the occluded image leads to 2.0\,\% accuracy, which is close the chance level.

Interestingly, $f$ in \cref{eq:objfun2} has local maxima corresponding to all scene motions, and motion computation involves searching for all maxima.
In comparison to alternative algorithms such as optical flow~\cite{horn-AI1981} and motion layers~\cite{wang-tip1994}, this approach is robust to occlusion and does not require the number of motions given.
The same computational process is used for both finding the motion and for reconstruction.
\section{Conclusions} \label{sec:conclusions}
The paper proposes an algorithm to reconstruct or de-occlude an object behind fragmented occlusion. It is an alternative to synthetic aperture techniques, and relies on motion between the object and the camera. 
%
Averaging registered images to create an integrated image and using sample variance as a contrast measure yields local maxima under certain conditions, as discussed in this paper. These maxima determine the object’s motion, and the resulting reconstructed object.
Using the integrated image and CLIP as a straightforward zero-shot classifier, achieves an impressive average 70.0\,\% accuracy for images with 90\,\% occlusion. This is significant because no other known classifiers perform well under such conditions. 
The algorithm is simple but provides useful results up to 90\,\% occlusion density and significantly improves state-of-the-art performance.
More complex motion models or robust estimation for generating the integrated image, have the potential to improve the results. 
Learning can be used throughout the process, but we did not use it yet to reduce the risk of hallucinations. We are currently investigating these directions.

The work focuses on psychological studies in dynamic perception.
The registration in the algorithm can be seen as a fixation on the object.  Interestingly, a human fixation on a point makes the rest of the retinal image blurry, a phenomenon known as Troxler's fading~\cite{alexander-reports2021}, but also observed in the integrated image. While the common explanation to the Troxler's fading is different, it would be interesting to see whether the algorithm provides a relevant explanation.
{
    \small
    \bibliographystyle{ieeenat_fullname}
    \bibliography{main, egbib}
}

\clearpage
\renewcommand{\thepage}{S\arabic{page}}
\setcounter{page}{1}
\clearpage
\setcounter{page}{1}
\maketitlesupplementary

\section{Text Labels}
\label{sec:labels}

For our robust recognition experiment in \cref{sec:recognition}, we used the following text labels coming from a set of 24 diverse outdoor categories $L$ = \{\texttt{goat}, \texttt{helicopter}, \texttt{swamp}, \texttt{drone}, \texttt{fireplace}, \texttt{train}, \texttt{camp}, \texttt{snowmobile}, \texttt{boat}, \texttt{cave}, \texttt{tent}, \texttt{rock},
\texttt{tree}, \texttt{lake}, \texttt{river},
\texttt{bridge}, \texttt{house}, \texttt{street}, \texttt{deer}, \texttt{fox}, \texttt{motorcycle}, \texttt{horse}, \texttt{car}, \texttt{airplane}\}.

\begin{figure}[h]
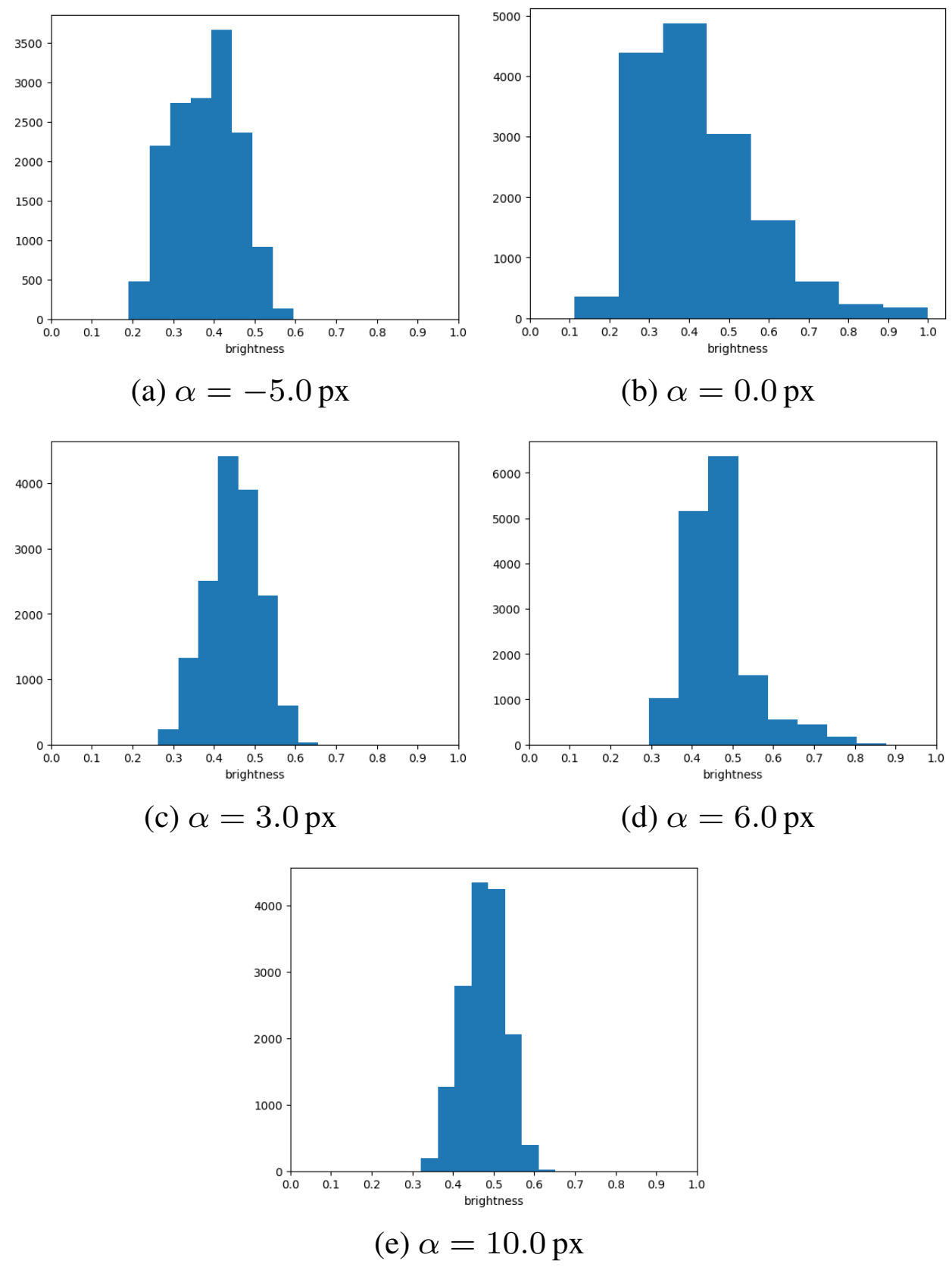

\centering
\begin{subfigure}[b]{.49\linewidth}
\includegraphics[width=\textwidth]{\figexperiment{020-tum-hgram-v_-05_00_}}
\caption{$\alpha = -5.0\,\text{px}$}
\end{subfigure}
\begin{subfigure}[b]{.49\linewidth}
\includegraphics[width=\textwidth]{\figexperiment{040-tum-hgram-v_+00_00_}}
\caption{$\alpha = 0.0\,\text{px}$}
\end{subfigure}\\[1ex]
\begin{subfigure}[b]{.49\linewidth}
\includegraphics[width=\textwidth]{\figexperiment{052-tum-hgram-v_+03_00_}}
\caption{$\alpha = 3.0\,\text{px}$}
\end{subfigure}
\begin{subfigure}[b]{.49\linewidth}
\includegraphics[width=\textwidth]{\figexperiment{064-tum-hgram-v_+06_00_}}
\caption{$\alpha = 6.0\,\text{px}$}
\end{subfigure}\\[1ex]
\begin{subfigure}[b]{.49\linewidth}
\includegraphics[width=\textwidth]{\figexperiment{080-tum-hgram-v_+10_00_}}
\caption{$\alpha = 10.0\,\text{px}$}
\end{subfigure}
\caption{Park sequence histograms.}
\label{fig:histopark}
\end{figure}

\section{Measures}
\label{sec:measures}

The mean-squared error (MSE) between two images X and Y is defined
\begin{equation}
\text{MSE}(X, Y) = \frac{1}{N} \sum_{i, j} (X(i, j) - Y(i, j))^2.
\end{equation}

The Pearson correlation coefficient between images $X$ and $Y$ is computed by vectorizing $X$ and $Y$, \ie $x = \vecop{X}$, $y = \vecop{Y}$ and then by computing their mean values $m_x$ and $m_y$. The coefficient is then written as
\begin{equation}
\mu_\text{P} = \frac{\sum(x - m_x)(y - m_y)}{\sqrt{\sum(x - m_x)^2 \sum(y - m_y)^2}}.
\end{equation}

\section{Histograms}
We selected two sequences from \cref{sec:experiments} to show the histograms from integral images.

\cref{fig:histopark} shows the histograms of the integrated images depicted in \cref{fig:park} - Park sequence. The histograms' spreads translate into the variances plotted in \cref{fig:func:park}. The integrated image at $\alpha = 0$ has largest variance (spread), the integrated image at $\alpha = 6.0$ has second-largest variance (spread) which relates to the person's motion.

\begin{figure}[h]
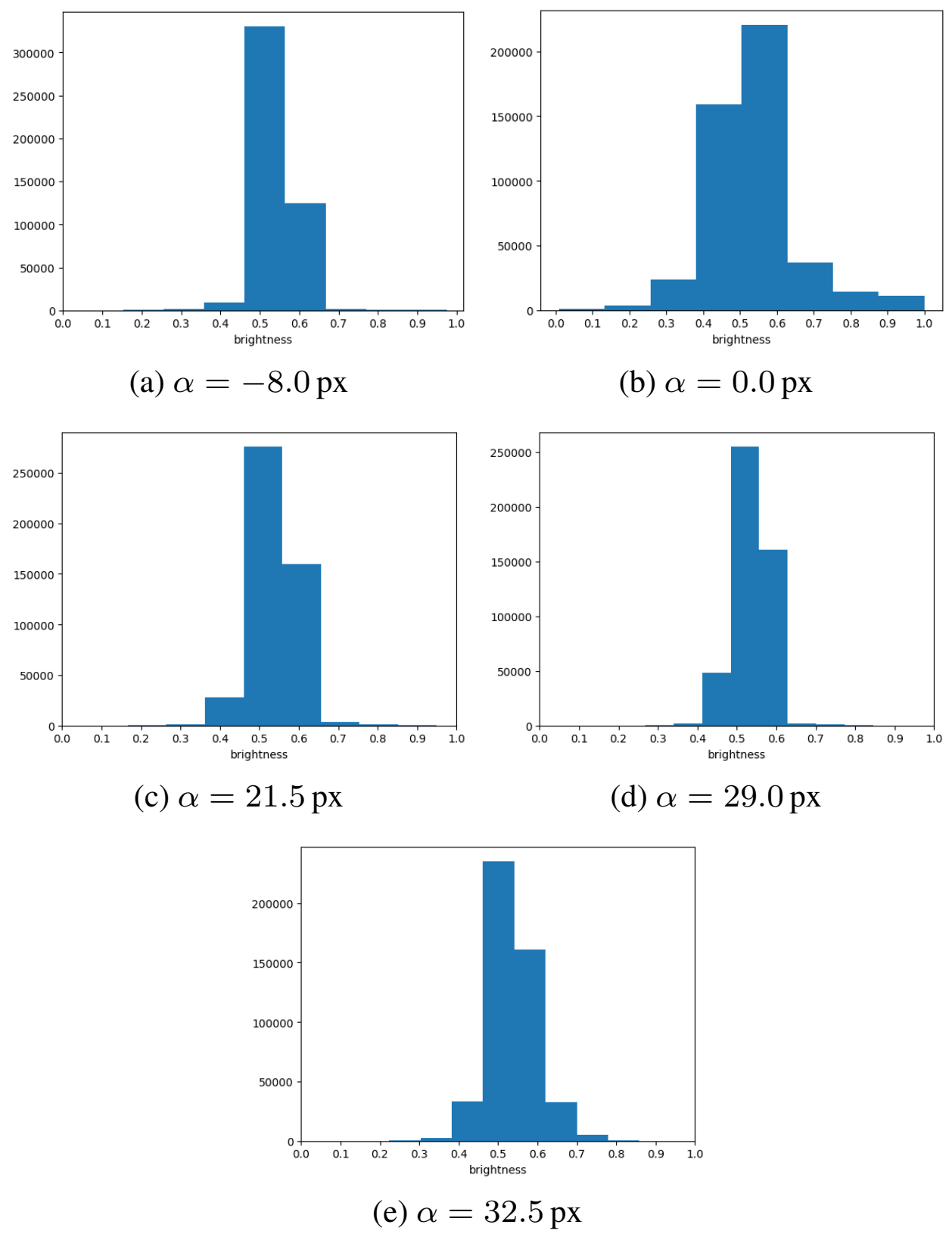

\centering
\begin{subfigure}[b]{.49\linewidth}
\includegraphics[width=\textwidth]{\figexperiment{004-sfence-grey-hgram-v_-08_00_}}
\caption{$\alpha = -8.0\,\text{px}$}
\end{subfigure}
\begin{subfigure}[b]{.49\linewidth}
\includegraphics[width=\textwidth]{\figexperiment{020-sfence-grey-hgram-v_+00_00_}}
\caption{$\alpha = 0.0\,\text{px}$}
\end{subfigure}\\[1ex]
\begin{subfigure}[b]{.49\linewidth}
\includegraphics[width=\textwidth]{\figexperiment{063-sfence-grey-hgram-v_+21_50_}}
\caption{$\alpha = 21.5\,\text{px}$}
\end{subfigure}
\begin{subfigure}[b]{.49\linewidth}
\includegraphics[width=\textwidth]{\figexperiment{078-sfence-grey-hgram-v_+29_00_}}
\caption{$\alpha = 29.0\,\text{px}$}
\end{subfigure}\\[1ex]
\begin{subfigure}[b]{.49\linewidth}
\includegraphics[width=\textwidth]{\figexperiment{085-sfence-grey-hgram-v_+32_50_}}
\caption{$\alpha = 32.5\,\text{px}$}
\end{subfigure}
\caption{Fence sequence histograms.}
\label{fig:histofence}
\end{figure}

Moreover, \cref{fig:histofence} shows the histograms of the integrated images depicted in \cref{fig:fence} - Fence sequence.

\begin{figure*}
\centering
\begin{subfigure}[b]{.24\linewidth}
\includegraphics[width=\textwidth]{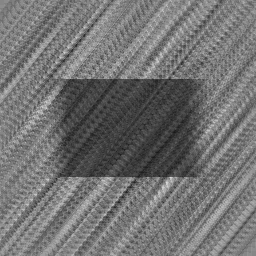}
\includegraphics[width=\textwidth]{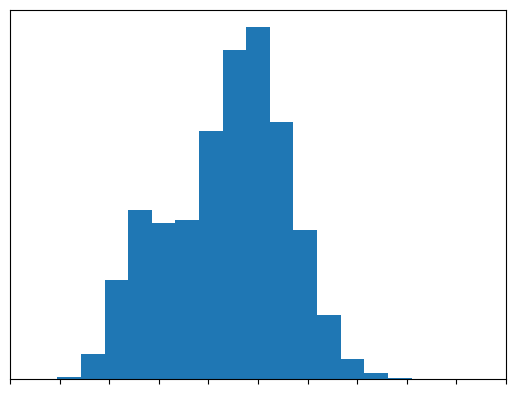}
\caption{$\alpha = -5.0\,\text{px}, \beta = 4.0\,\text{px}$}
\end{subfigure}
\begin{subfigure}[b]{.24\linewidth}
\includegraphics[width=\textwidth]{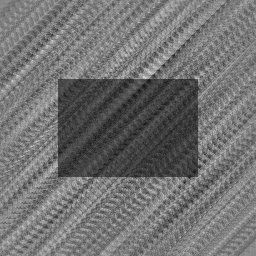}
\includegraphics[width=\textwidth]{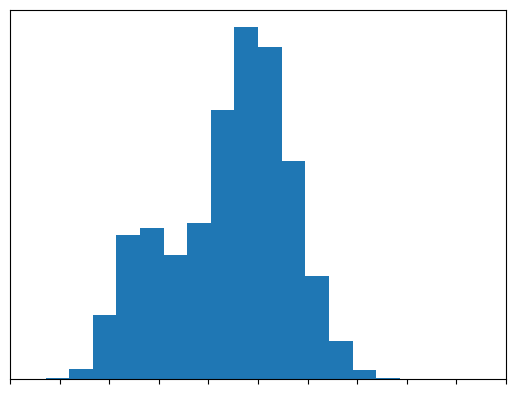}
\caption{$\alpha = -5.0\,\text{px}, \beta = 5.0\,\text{px}$}
\end{subfigure}
\begin{subfigure}[b]{.24\linewidth}
\includegraphics[width=\textwidth]{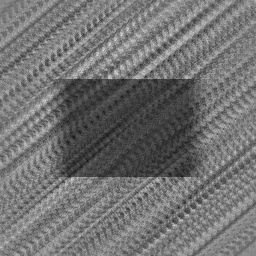}
\includegraphics[width=\textwidth]{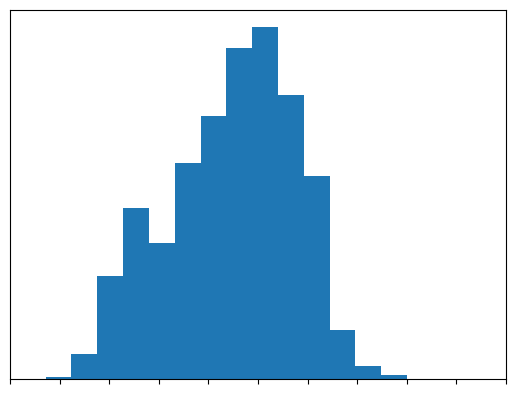}
\caption{$\alpha = -5.0\,\text{px}, \beta = 6.0\,\text{px}$}
\end{subfigure}
\begin{subfigure}[b]{.24\linewidth}
\includegraphics[width=\textwidth]{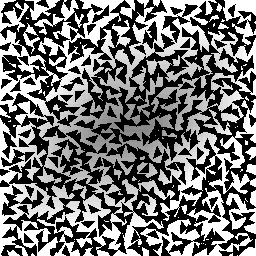}
\includegraphics[width=\textwidth]{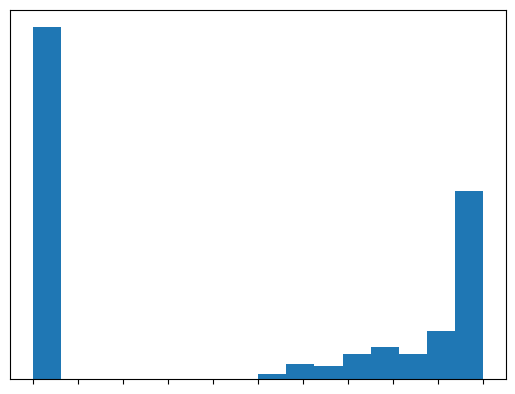}
\caption{$\alpha = 0.0\,\text{px}, \beta = 0.0\,\text{px}$}
\end{subfigure}
\caption{Evaluation of different motions. Top: The variance of (a) and (c) is larger or equal to (b), although (b) shows higher contrast than (a) and (c) and is the integrated image of the correct box's motion. Bottom: This is verified by the histogram spreads. (d) illustrates the static occluder. The histogram spread for (d) is maximal (255) due to the chosen brightness values for the background and occluder.}
\label{fig:boxresults}
\end{figure*}

\section{Variance: Failure Case} \label{sec:failurecase}
Motion computation in \cref{eq:learning} successfully works for all 300 synthetic image sequences in \cref{sec:experiments}. However, when applied to real image sequences, the integrated sample variance becomes significantly larger than the variance associated with the occluded object. Consequently, the function $f$ fails to attain its maximum value. This raises the question of the conditions under which motion computation may fail.

\begin{figure}
\centering
\begin{subfigure}[b]{.32\linewidth}
\includegraphics[width=\textwidth]{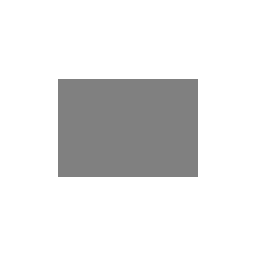}
\caption{$\bar{I}_r$}
\end{subfigure}
\begin{subfigure}[b]{.32\linewidth}
\includegraphics[width=\textwidth]{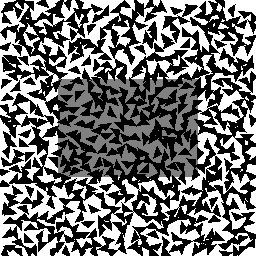}
\caption{$I_r$}
\end{subfigure}
\begin{subfigure}[b]{.32\linewidth}
\includegraphics[width=\linewidth]{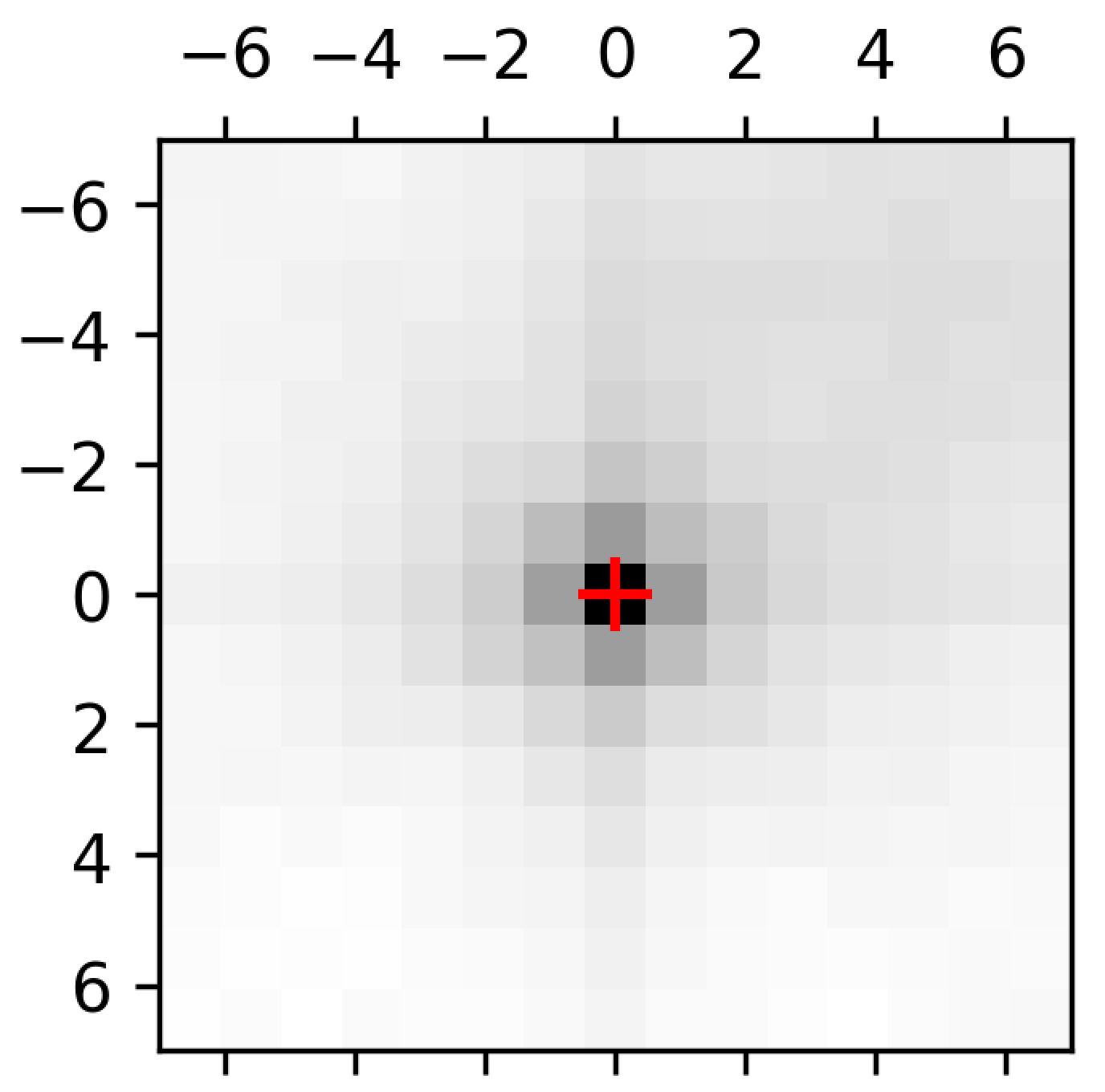}
\caption{$f$}
\end{subfigure}
\caption{Box experiment: (a) Ground truth reference image. (b) Sequence's reference image at reference $r=11$. (c) 2D visualization of $f$ - $\beta$ (abscissa) \vs $\alpha$ (ordinate): Colors encode the variance. The motion at $(\alpha~\beta)^\top = \mathbf{0}$ is found by the algorithm (\textcolor{red}{+}). $f$ attains no maximum for the box's motion $\rowvec{-5}{5}^\top$.}
\label{fig:box}
\end{figure}

To study such conditions, we generated a sequence of 21 synthetic images, showing an artificial box under occlusion density $\rho = 49.3\,\%$ (see \cref{fig:box}). The box's brightness value is mid-gray (128), the background white (255), and the occluder black (0). \cref{fig:boxresults} illustrates the results. Motion computation fails under these conditions, because the range of the average brightness values within the integrated image includes the brightness value of the box object. This can be nicely seen by the histograms' spreads for the different motion parameters $\alpha$ and $\beta$ in \cref{eq:theta}.

\begin{figure}
\centering
\begin{subfigure}[b]{.7\linewidth}
\includegraphics[width=\textwidth]{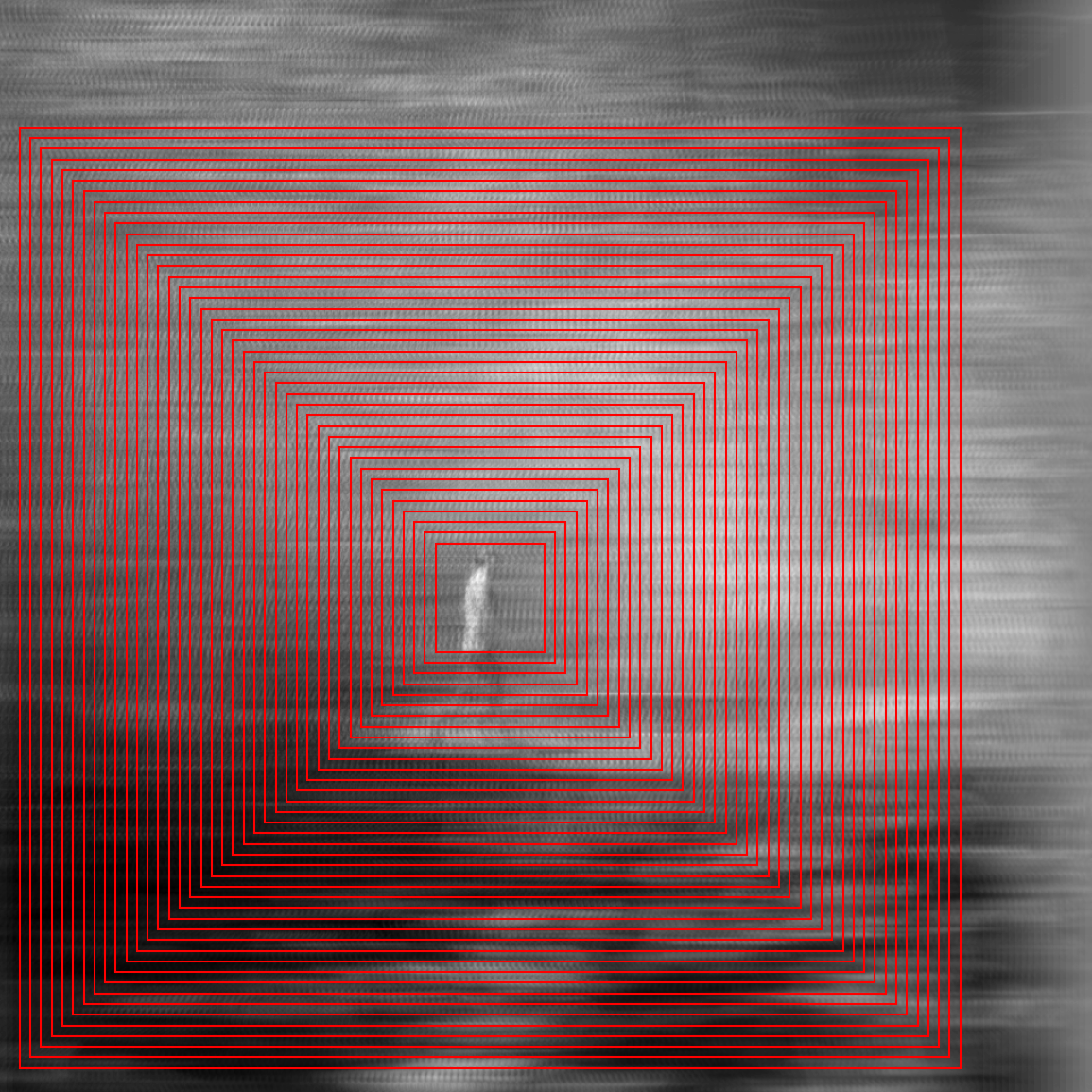}
\caption{Fixation windows}
\label{fig:fixwindow:windows}
\end{subfigure}\\[1ex]
\begin{subfigure}[b]{.9\linewidth}
\includegraphics[width=\textwidth]{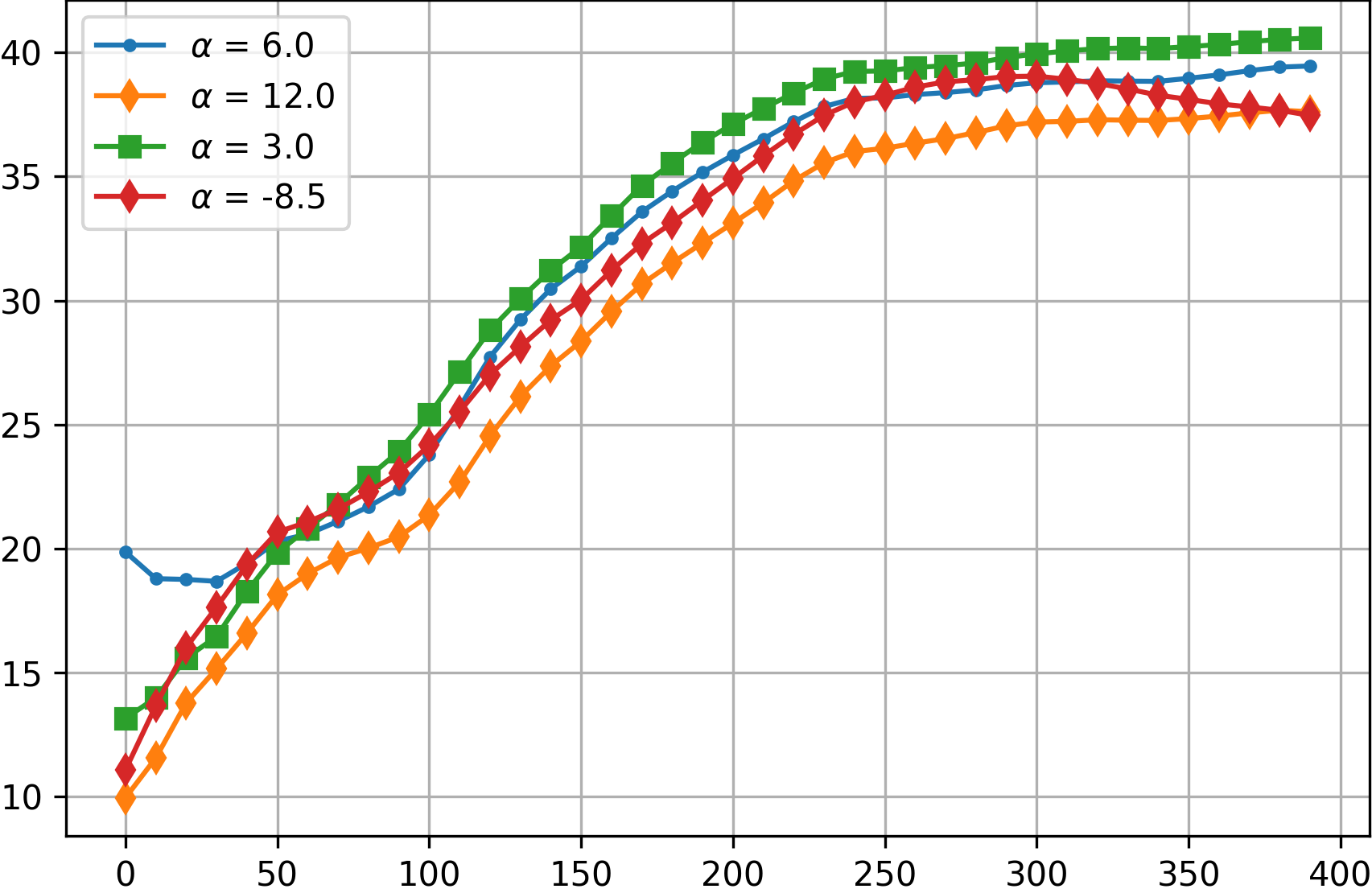}
\caption{Sample variance \vs window diagonal (px)}
\label{fig:fixwindow:eval}
\end{subfigure}
\caption{Evaluation of the fixation window size: (a) Windows of varying size. (b) Sample variance computed in the integrated image for the different windows. Note that the sample variance for the correct motion is significantly higher than for the incorrect motions when the fixation window closely catches the object.}
\label{fig:fixwindow}
\end{figure}

We conducted another experiment with the Park sequence, computing the sample variance for different fixation windows. These windows vary in size and capture the object or parts of it. \cref{fig:fixwindow} shows the results of this experiment. The sample variance is similar across window sizes for incorrect object motions. However, for small windows with larger object regions, the sample variance suddenly increases for the correct motion $\alpha = 6.0$.

We conclude that motion computation fails if the object brightness values are a subset of the integrated image. This is because there are no distinctive brightness values to increase variance, so $f$ does not attain a maximum for the correct motion. This is a fundamental limitation of variance in our algorithm. This may also explain Vaish~\etal’s observation that shape-from-focus (SfF) performs better. SfF exploits texture by incorporating image gradients instead of brightness values, which leads to better results~\cite{vaish-cvpr2006}.

\section{Multiple Objects} \label{sec:multipleobjects}
This experiment tests the algorithm’s behavior with six cartoon styles: Ape, Man, Crabby, Biker, Rake, and Fish. Unlike the synthetic experiment with a single Ape cartoon in \cref{sec:experiments}, a fixed occlusion pattern is used for 90 generated sequences (15 per object and level).

\begin{figure}
\centering
\begin{subfigure}[b]{.19\linewidth}
\includegraphics[width=\textwidth]{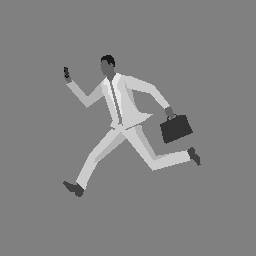}
\includegraphics[width=\textwidth]{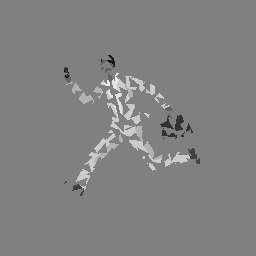}
\includegraphics[width=\textwidth]{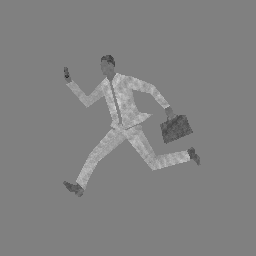}
\caption{48\,\%}
\label{fig:cartoon15man}
\end{subfigure}
\begin{subfigure}[b]{.19\linewidth}
\includegraphics[width=\textwidth]{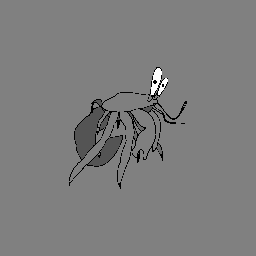}
\includegraphics[width=\textwidth]{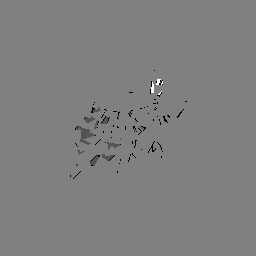}
\includegraphics[width=\textwidth]{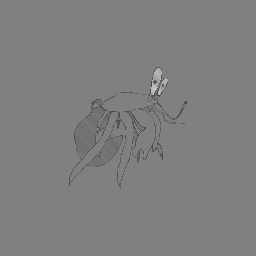}
\caption{67.8\,\%}
\label{fig:cartoon15crabby}
\end{subfigure}
\begin{subfigure}[b]{.19\linewidth}
\includegraphics[width=\textwidth]{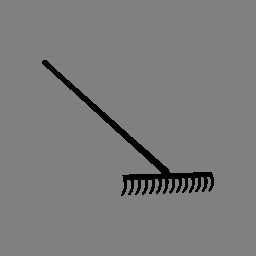}
\includegraphics[width=\textwidth]{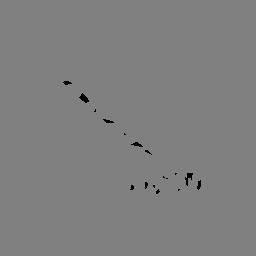}
\includegraphics[width=\textwidth]{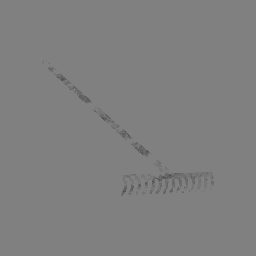}
\caption{87.8\,\%}
\label{fig:cartoon15rake}
\end{subfigure}
\begin{subfigure}[b]{.19\linewidth}
\includegraphics[width=\textwidth]{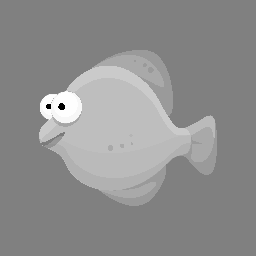}
\includegraphics[width=\textwidth]{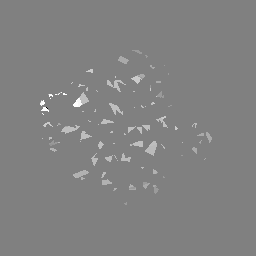}
\includegraphics[width=\textwidth]{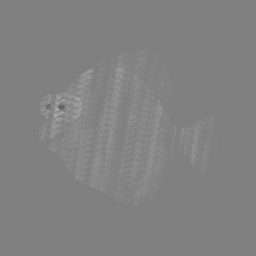}
\caption{90.1\,\%}
\label{fig:cartoon15fish}
\end{subfigure}
\begin{subfigure}[b]{.19\linewidth}
\includegraphics[width=\textwidth]{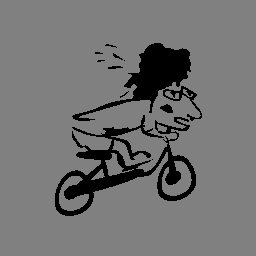}
\includegraphics[width=\textwidth]{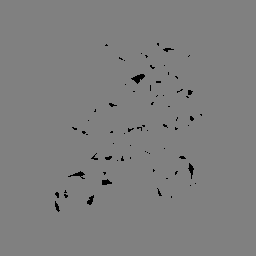}
\includegraphics[width=\textwidth]{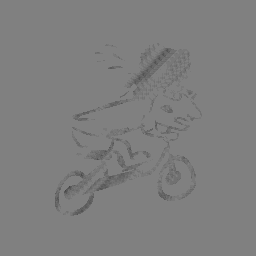}
\caption{90.8\,\%}
\label{fig:cartoon15biker}
\end{subfigure}\\[1ex]
\begin{subfigure}[b]{.49\linewidth}
\includegraphics[width=\textwidth]{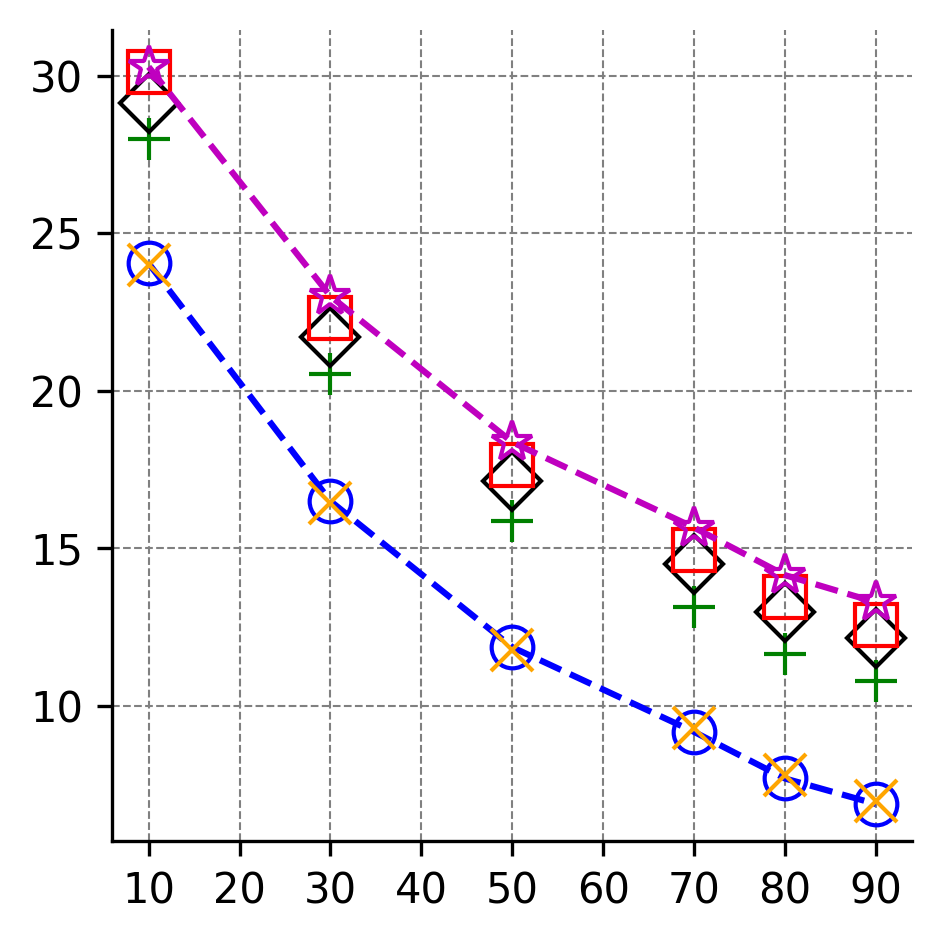}
\caption{$\mu_\text{PSNR}$}
\label{fig:cartoon15psnr}
\end{subfigure}
\begin{subfigure}[b]{.49\linewidth}
\includegraphics[width=\textwidth]{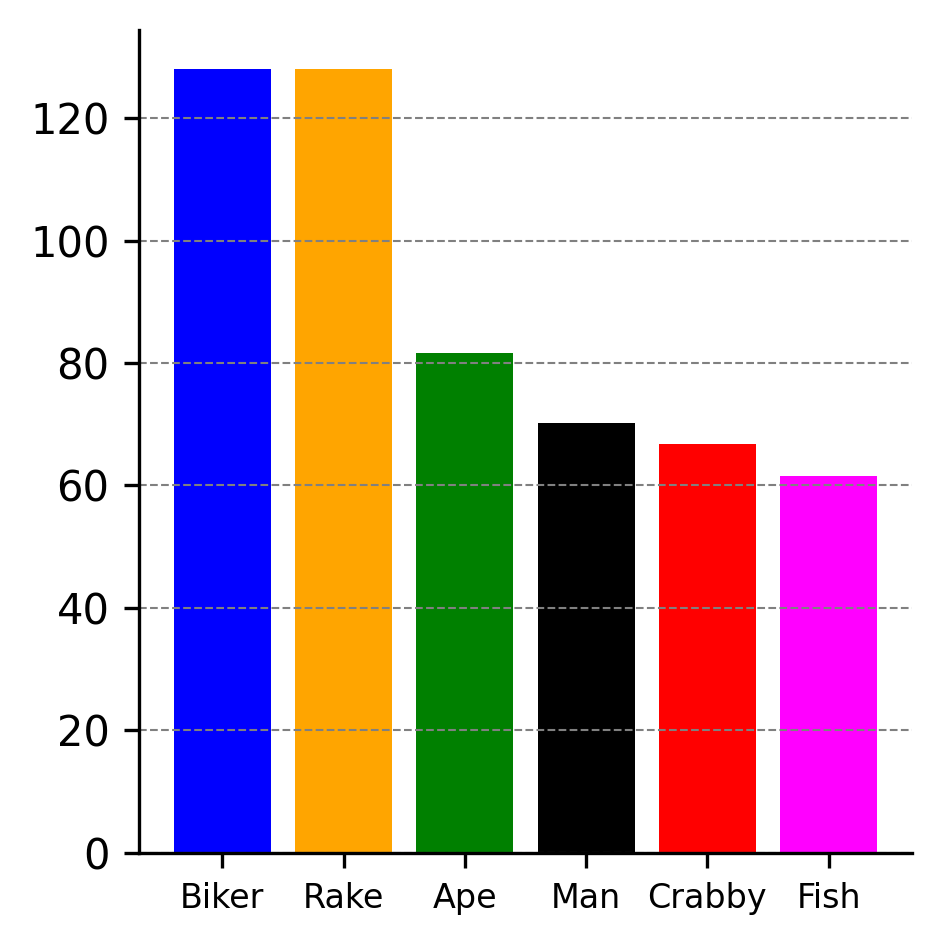}
\caption{Color deviation}
\label{fig:cartoon15dev}
\end{subfigure}
\caption{Experiment with multiple objects: (a) - (e) For varying occlusion densities, the images show the ground truth $\bar{I}_r$ (top), raw data $I_r$ (middle) and reconstructed images $\hat{I}_r$ (bottom). (f) Plot of PSNR (dB) \vs occlusion density. Colors are the same as in (g) Deviation (RMSE) between object and occluder intensities. Colors encode the six objects.}
\label{fig:cartoon15}
\end{figure}

In this experiment, we also use the peak signal-to-noise ratio (PSNR), a widely used image quality measure, in addition to the provided metrics:
\begin{equation}
\mu_\text{PSNR} = 10 \log \frac{255^2}{\text{MSE}(M_r \hat{I}_r, M_r \bar{I}_r)}.
\end{equation}

We found that the algorithm successfully reconstructs cartoon objects in all 90 sequences (see \crefrange{fig:cartoon15man}{fig:cartoon15fish}). Evaluation of $\mu_\text{PSNR}$ shows no correlation between intensity or shape and reconstruction quality (see \cref{fig:cartoon15psnr}). The reconstruction mainly depends on the deviation of object brightness from the occluder brightness (see additionally \cref{fig:cartoon15dev}). This deviation is computed as RMSE = $\sqrt{\text{MSE}}$ between object and occluder intensities. Smaller deviations result in a higher $\mu_\text{PSNR}$ because occluder intensities have a lesser impact on the averaging, which demonstrates the well known sensitivity of averaging (or the mean) to outliers.

\section{Videos, Data and Code} \label{sec:videos}
We provide the following four videos to aid readers in understanding the algorithm in \cref{sec:method}:
\textbf{park-sweep.mp4}, \textbf{boreal-sweep.mp4}, \textbf{pan-sweep.mp4}, \textbf{fence-sweep.mp4}. The videos show the integrated images and illustrate a sweep through the motion parameter $\alpha$ in \cref{eq:theta}.

For each real sequence and the Box experiment (see \cref{sec:failurecase}), we additionally provide two zip files: one with raw input image sequences and the other with intermediate results (histograms, integrated images). We do not provide data for the synthetic experiment and the multiple objects in \cref{sec:multipleobjects}, because adding this data exceeds the supplemental material size limit.
Upon acceptance, we will provide this data and the algorithm implementation to the reader.

\end{document}